\RequirePackage{fix-cm}
\documentclass[twocolumn]{svjour3}          
\smartqed  
\usepackage{graphicx}
\usepackage{amssymb}
\usepackage[cmex10]{amsmath}
\usepackage{array}
\usepackage{url}
\usepackage{color}
\usepackage[tight,footnotesize]{subfigure}
\usepackage{natbib}
\usepackage{hyphenat}
\usepackage[pdftex,
CJKbookmarks=true, bookmarksnumbered=true, bookmarksopen=true,
colorlinks=true,
pdfborder=001,
citecolor=blue, linkcolor=blue, anchorcolor=red, urlcolor=black
]{hyperref}
\usepackage{bbm,dsfont}
\usepackage{indentfirst} 
\usepackage{booktabs}

\usepackage[dvipsnames]{xcolor} 
\usepackage{caption}
\usepackage{subcaption}
\usepackage{multirow}
\usepackage{tabularx}
\usepackage{adjustbox}
\usepackage{pifont}
\usepackage[ruled,vlined]{algorithm2e}
\usepackage[skip=2pt]{caption}
\usepackage{marvosym}

\usepackage{bm}
\newcommand{\tsb}{\textsubscript}

\journalname{International Journal of Computer Vision}
\renewcommand{\arraystretch}{1.3}

\definecolor{darkmag}{rgb}{0.55,0,0.55}
\begin{document}
\begin{sloppypar}

\title{EKPC: Elastic Knowledge Preservation and Compensation for Class-Incremental Learning}

\titlerunning{Short form of title}        


\author{Huaijie Wang$^{*}$ \and
        De Cheng$^{*}$ \thanks{$^{\ast}$ Equal contribution.}\and
        Lingfeng He\and
        Yan Li \and
        Jie Li  \and
        Nannan Wang \textsuperscript{\Letter}\and
        Xinbo Gao \textsuperscript{\Letter} 
}

\authorrunning{Short form of author list} 

\institute{
Huaijie Wang$^{*}$ Equal contribution\at
Xidian University, Xi'an 710071, China \\
\email{\href{huaijie_wang@stu.xidian.edu.cn}{huaijie\_wang@stu.xidian.edu.cn}}
\and
De Cheng$^{*}$  Equal contribution\at
Xidian University, Xi'an 710071, China \\
\email{\href{dcheng@xidian.edu.cn}{dcheng@xidian.edu.cn}} 
\and
Lingfeng He \at
Xidian University, Xi'an 710071, China \\
\email{\href{lfhe@stu.xidian.edu.cn}{lfhe@stu.xidian.edu.cn}} 
\and
Yan Li \at
Northwestern Polytechnical University, Xi'an 710129, China\\
\email{\href{yanli.ly.cs@gmail.com}{yanli.ly.cs@gmail.com}} 
\and
Jie Li\at
Xidian University, Xi'an 710071, China \\
\email{\href{leejie@mail.xidian.edu.cn}{leejie@mail.xidian.edu.cn}}
\and
Xinbo Gao\textsuperscript{\Letter}, Corresponding Author\at
Chongqing University of Posts and Telecommunications, Chongqing 400065, China \\
\email{\href{gaoxb@cqupt.edu.cn}{gaoxb@cqupt.edu.cn}}
\and
Nannan Wang\textsuperscript{\Letter}, Corresponding Author \at
Xidian University, Xi'an 710071, China \\
\email{\href{nnwang@xidian.edu.cn}{nnwang@xidian.edu.cn}}
}


\maketitle
\begin{abstract}

Class-Incremental Learning (CIL) aims to enable AI models to continuously learn from sequentially arriving data of different classes over time while retaining previously acquired knowledge. Recently, Parameter-Efficient Fine-Tuning (PEFT) methods, like prompt pool-based approaches and adapter tuning, have shown great attraction in CIL. However, these methods either introduce additional parameters that increase memory usage, or rely on rigid regularization techniques which reduce forgetting but compromise model flexibility. To overcome these limitations, we propose the Elastic Knowledge Preservation and Compensation (EKPC) method, integrating Importance-aware Parameter Regularization (IPR) and Trainable Semantic Drift Compensation (TSDC) for CIL. Specifically, the IPR method assesses the sensitivity of network parameters to prior tasks using a novel parameter-importance algorithm. It then selectively constrains updates within the shared adapter according to these importance values, thereby preserving previously acquired knowledge while maintaining the model’s flexibility. However, it still exhibits slight semantic differences in previous knowledge to accommodate new incremental tasks, leading to decision boundaries confusion in classifier. To eliminate this confusion, TSDC trains a unified classifier by compensating prototypes with trainable semantic drift. Extensive experiments on five CIL benchmarks demonstrate the effectiveness of the proposed method, showing superior performances to existing state-of-the-art methods. 
\end{abstract}
\section{Introduction}
Continual learning (CL) \cite{belouadah2021comprehensive, de2021continual, masana2022class, zhang2024center} is a machine learning approach where a model learns new tasks or knowledge sequentially over time while retaining what it has previously learned. This approach is crucial in dynamic environments where data arrives in stages or where models need to handle an expanding range of tasks, such as in robotics, personalized recommendations, or autonomous systems. The primary objective of CL is to allow a model to adapt to new information without ``catastrophic forgetting \cite{CatastrophicInterference1989McCloskey, ConnectionistModels1990Ratcliff}"—a phenomenon where learning new information disrupts or erases existing knowledge. CL typically includes two main scenarios: Task-Incremental Learning (TIL) \cite{soutif2021importance} and Class-Incremental Learning (CIL) \cite{li2017learning}. In this work, we focus on the more challenging CIL setting, where task identities are unknown during inference, making it harder to distinguish between old and new classes.\par
\begin{figure}[t]
\centering
\includegraphics[width=\linewidth]{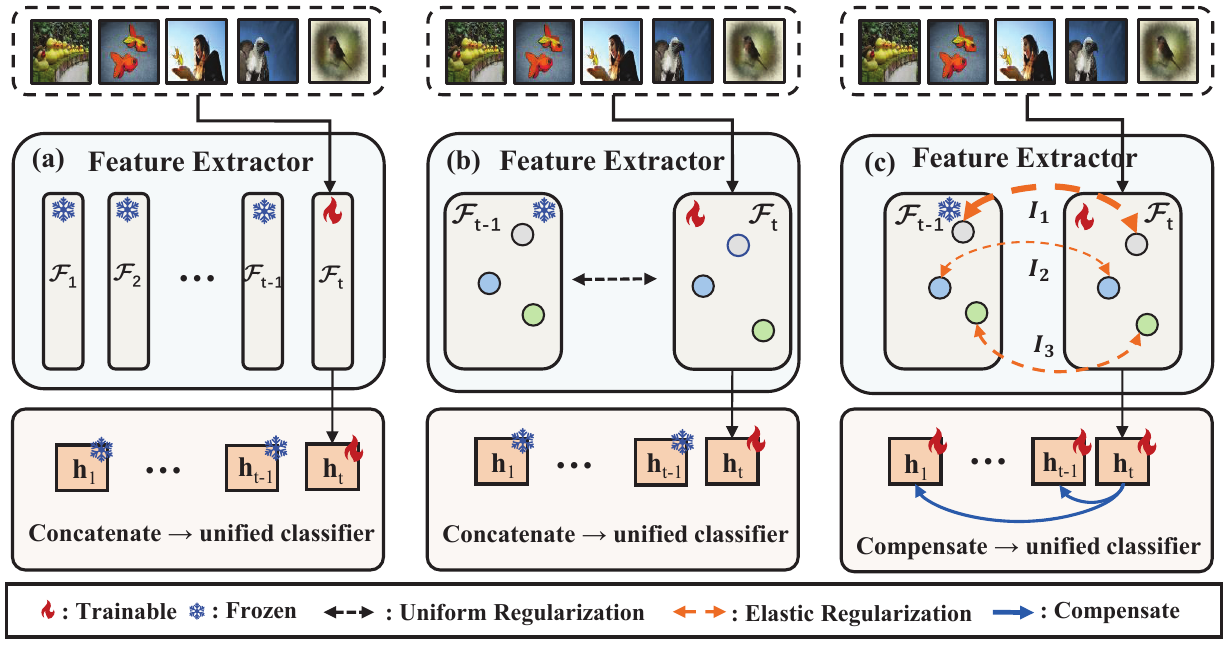}
\caption{(a) illustrates the expandable parameters approach, where each task uses a separate feature extractor. (b) depicts a shared feature extractor with non-importance-based regularization. (c) presents our EKPC method, employing elastic parameter regularization based on importance. Below each figure, in (a) and (b), classifiers are concatenated to form a unified model, while in Fig 1(c), a unified classifier is trained via a compensation mechanism.}
\label{Introduction}
\end{figure}
Recent progress in CIL has largely focused on fine-tuning foundation models, taking advantage of the generalization power of pre-trained models \cite{dosovitskiy2020image}, which are often used as feature extractors in CIL tasks. Type of prompt pool-based methods  \cite{wang2022learning, wang2022dualprompt, CODAPromptCOntinual2023Smith}, maintain a collection of prompts for all tasks and select the most suitable prompt for each class. Another approach, adapter tuning methods\cite{gao2024beyond,tan2024semantically,zhou2024revisiting}, addresses continual learning by adding either expandable adapters or shared adapters. While expandable adapters increase extra parameter counts(Fig.\ref{Introduction}(a)), shared adapters often suffer from more forgetting, as parameters learned from previous tasks are updated when new tasks are introduced. To address forgetting, many methods use regularization \cite{kirkpatrick2017overcoming, zenke2017continual, aljundi2018memory} to reduce changes to parameters across tasks, but this typically involves applying uniform constraints across all parameters, which can reduce model flexibility and adaptability (Fig.\ref{Introduction}(b)). This presents a key challenge: how to achieve effective anti-forgetting without adding extra parameters, while maintaining stability and preserving model plasticity during training.\par
To address these challenges in CIL, we introduce the Elastic Knowledge Preservation and Compensation (EKPC) method(Fig.\ref{Introduction}(c)), which comprises two main components: 1) the Importance-aware Parameter Regularization (IPR) for the shared adapter in the feature extractor, and 2) the Trainable Semantic Drift Compensation (TSDC) for training a unified classifier. The IPR module first calculates the importance of the shared adapter's network parameters to the previous task through both global and local aspects. Then, it applies adaptive regularization weights to the shared adapter based on their importance, minimizing the changes in key parameters between tasks and relax the restrictions on other parameter learning new tasks. This approach not only explains the sensitivity of networks parameters to previous tasks, but preserves past knowledge in feature space without increasing parameter count and maintains flexibility for learning new knowledge. However, there are still subtle semantic differences across tasks, causing the decision boundaries confusion within the classifier. To address this, TSDC is introduced to adjust previous class prototypes by estimating and regularizing their drift, thus compensating for cross-task discrepancies in class representations. This compensation mitigates the decision boundary confusion. Through IPR and TSDC, EKPC achieves a strong balance between stability and plasticity, resulting in a more robust and accurate model. \par
IPR quantifies the global importance of network parameters by optimizing the classifier’s performance—maximizing its score while minimizing variance—across the various channels of the final output feature. Recognizing that adapters at different depths contribute heterogeneously to the output, IPR further assesses the local importance of each adapter module by analyzing the sensitivity of the output to targeted perturbations applied to it. When regularizing, unlike methods that apply uniform constraints across all parameters, IPR adaptively constrain parameters based on their importance. This approach preserves model stability by focusing on the primary components of model parameters while without sacrificing too much model flexibility.\par
While IPR adaptively preserves past knowledge for the shared adapter, slight semantic differences still exist in the feature space across tasks, leading to confusion of decision boundaries within classifier. Since addressing this issue requires managing the evolving decision boundaries as new tasks are introduced, static estimation method may has large deviations, which does not prevent boundary confusion over time. To address this challenge, TSDC introduces a trainable method. When estimating the drift by current data, TSDC regularizes drift estimates toward the zero subspace, improving the precision of prototype adjustments benefiting a unified classifier training. This strategy minimizes task-related boundary confusion, ultimately leading to a more robust and accurate unified classifier.
The main contributions can be summarized as follows:
\begin{itemize}
\item We introduce a importance-aware parameter regularization module to retain past knowledge elastically. It assesses global importance by formulating channel-wise optimization for quantifying all adapters' contribution and evaluates local importance by analyzing output perturbations for capturing the specific adapter's contribution. This approach preserves stability while maintaining model plasticity.
\item We introduce a trainable semantic drift compensation module to mitigate prototype drift. During training, regularization reduces the deviation of estimated semantic drift from the zero subspace. This stabilization enhances decision boundary consistency and ensures a robust, unified classifier.
\item Extensive experimental results demonstrate the effectiveness of  of our proposed method in mitigating forgetting across five class incremental learning benchmarks with diverse experiment settings, and our approach achieves superior performances to state-of-the-art methods.
\end{itemize}
\section{Related Work}
\subsection{Class-Incremental Learning}
Class-Incremental Learning (CIL) aims to enable models to continuously learn new classes while retaining knowledge of previous ones, addressing the fundamental challenge of balancing stability (preventing catastrophic forgetting) and plasticity (adapting to new tasks) \cite{zhou2023deep}. Existing methodologies can be broadly categorized into three paradigms, each with distinct mechanisms and trade-offs:
regularization-based \cite{kirkpatrick2017overcoming, zenke2017continual, xiang2022coarse, zhou2023hierarchical}, replay-based \cite{bang2021rainbow, chaudhry2018riemannian, rebuffi2017icarl}, and  structural adjustment-based approaches \cite{serra2018overcoming, mallya2018packnet, mallya2018piggyback, liang2024inflora, yu2024boosting}.
Regularization-based methods constrain parameter updates to preserve critical information from prior tasks, such as knowledge distillation \cite{li2017learning, hou2019learning, tao2020few, helingfeng}, parameter regularization \cite{chaudhry2018riemannian, kirkpatrick2017overcoming, zenke2017continual} use a regularization method to reduce the change of model to mitigates forgetting.
Replay-based methods retain or regenerate samples from past tasks to stabilize training. They explicitly address forgetting by replaying historical data alongside new task data. Subcategories include: Direct Replay \cite{bang2021rainbow} and Generative \cite{shin2017continual} reduce forgetting by performing joint training on new tasks while retaining or generating previous class data, preventing over-fitting to new tasks.
Structural adjustment-based methods dynamically modify network architectures to isolate task-specific knowledge or expand model capacity. Key variants include: neuron expansion \cite{ostapenko2019learning} and backbone expansion \cite{yan2021dynamically}, learn new tasks with expandable parameters while freezing the previous model to reduce forgetting.
\begin{figure*}[htbp]
\centering
\includegraphics[width=\linewidth]{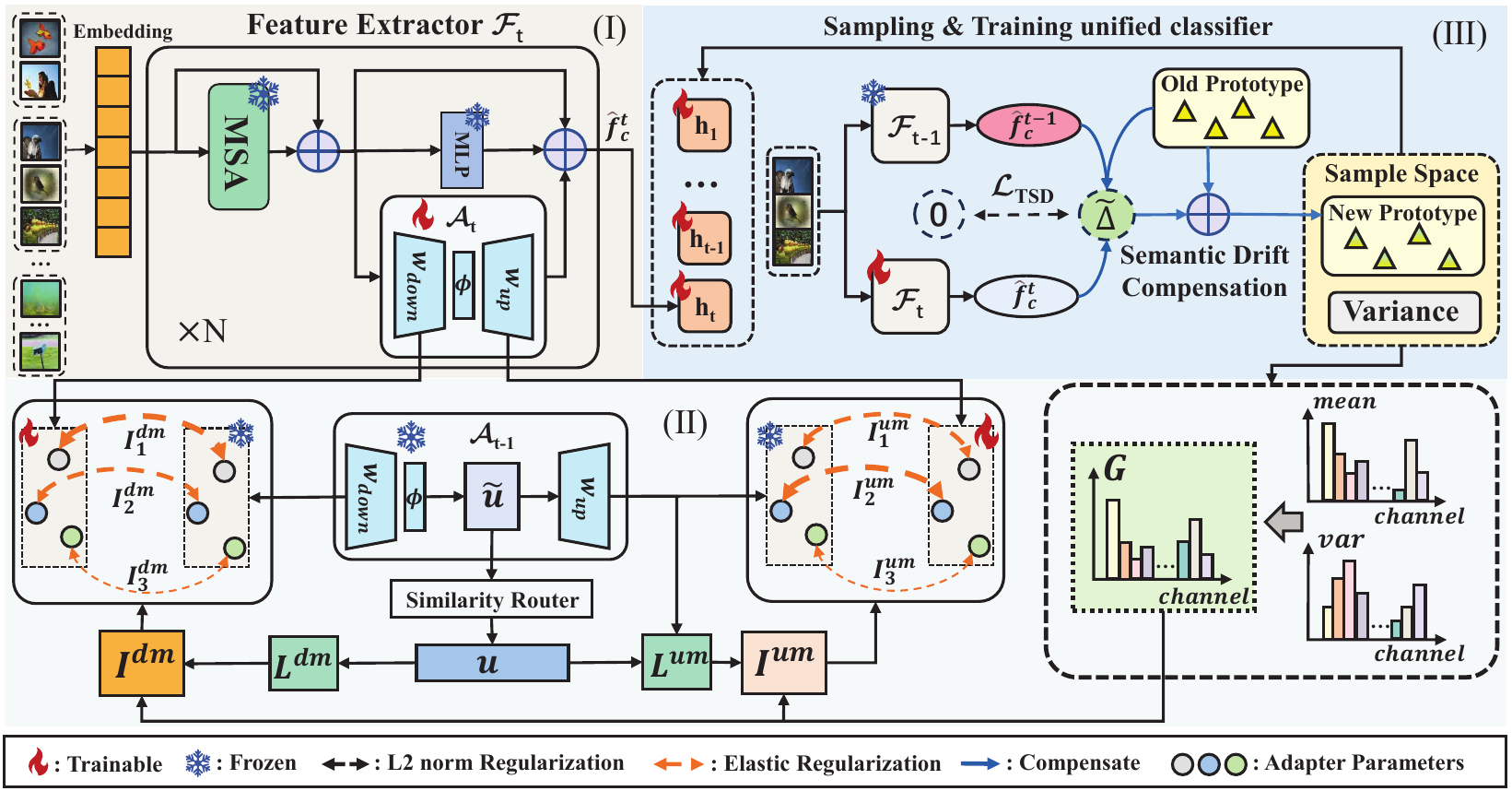}
\caption{Overview of our proposed framework. (\uppercase\expandafter{\romannumeral1}) a backbone network with shared adapter tuning, (\uppercase\expandafter{\romannumeral2}) Importance-Aware Parameter Regularization (IPR) that dynamically balances stability and plasticity based on task importance, and (\uppercase\expandafter{\romannumeral3}) Trainable Semantic Drift Compensation (TSDC) to reduce decision boundary ambiguity in the unified classifier by estimated and regularized semantic drift.}
\label{framework}
\end{figure*}
\subsection{Parameter-Efficient Tuning}
Parameter-Efficient Fine-Tuning (PEFT) has emerged as a paradigm shift in adapting large-scale pre-trained models to downstream tasks while minimizing computational and memory overhead. By updating only a small subset of model parameters, PEFT preserves the generalizable knowledge encoded in pre-trained weights while enabling task-specific adaptation. Initially, this approach showed promising results in transfer learning tasks in NLP \cite{houlsby2019parameter, lester2021power, li2021prefix, hu2021lora}. For example, AdaptFormer \cite{chen2022adaptformer} integrates lightweight modules after MLP layers, outperforming full model fine-tuning on action recognition. SSF \cite{lian2022scaling} achieves competitive results using fewer parameters. Inspired by prompt-based language model strategies, VPT \cite{jia2022visual} is a method that fine-tunes the visual models with minimal parameter increases. Prompt-based methods also enhance performance in vision-language models \cite{radford2021learning, zhou2022learning, zhou2022conditional, zhang2022pointclip, xuzhipeng}.\par 
\subsection{Parameter-Efficient Tuning in CIL}
Recently, CIL with pre-trained vision transformer models \cite{ermis2022memory, CODAPromptCOntinual2023Smith, wang2022dualprompt, wang2022learning} have demonstrated excellent performance. However, integrating PEFT with Class-Incremental Learning (CIL) introduces unique challenges: preserving historical knowledge while adapting to new classes with minimal parameter updates. Recent advances leverage pre-trained vision transformers (ViTs) as backbones, exploiting their transferable representations and modular architectures. Some approaches completely fine-tune pre-trained models, such as \cite{boschini2022transfer, zhang2023slca}, but these methods are often very time-intensive. Other strategies integrate PEFT (Parameter-Efficient Fine-Tuning) methods within continual learning. Prompt pool-based approaches \cite{wang2022learning, CODAPromptCOntinual2023Smith, wang2024hierarchical, zhang2023slca} maintain a collection of prompts for all tasks, selecting the most suitable prompts for each specific task. Meanwhile, Adam-adapter method \cite{zhou2023revisiting} proposes an adapted based approach on CIL. SSIAT \cite{tan2024semantically} uses shared adapter without regularization, having sufficient plasticity to learn new tasks. Other methods \cite{gao2024beyond, liang2024inflora, ExpandableSubspace2024Zhou} use expandable architectures with adapters, aiming to mitigate catastrophic forgetting by applying regularization terms or shift based adjustments. As these methods also have limitation of growing parameters or losing plasticity due to regularization, we propose EKPC to address this challenge.
\section{Methodology}
\textbf{Problem Definition}: 
In a continual learning setting, a model is trained across $T$ stages with the sequentially arriving datasets
$\{\mathcal{D}^{1},\dots, \mathcal{D}^{T}\}$. $\mathcal{D}^{t}=\{ (\mathbf{x}_{j}^{t},y_{j}^{t}) \}_{j=1}^{|\mathcal{D}^{t}|}$ denotes the dataset of the $t$-th task, where $\mathbf{x}_{j}^{t}$ is the $j$-th inputs of $t$-th task with label $y_{j}^{t}$, and $|\mathcal{D}^{t}|$ represents the number of samples for the $t$-th task. $\mathcal{Y}^{t}$ is the label space of the $t$-th training set. For $\forall t,t'$, when $t\ne t'$, $\mathcal{Y}^{t} \cap\mathcal{Y}^{t'}=\varnothing$ means that different tasks share disjoint classes. The goal of continual learning is to train a model across $T$ tasks sequentially and perform well on all the learned classes $\{\mathcal{Y}^{1},\mathcal{Y}^{2},\dots,\mathcal{Y}^{t}\}$. \par
\noindent \textbf{Overall framework}:  
We follow SSIAT \cite{tan2024semantically} that fine-tunes the backbone of Vision Transformer(ViT) \cite{dosovitskiy2020image} with shared adapters. Considering a model $\mathbf{h}(\mathcal{F}(\cdot))$, $\mathcal{F}(\cdot)$ denotes the feature extractor, where the ViT parameters are frozen and only the adapter parameters are tunable. $\mathbf{h}(\cdot)$ represents the classifier, which is also trainable. We primarily 
focus on tackling the challenging class-incremental learning, where previous data is unavailable during model training, and the task ID is unknown during inference.\par
The overall framework of our EKPC is illustrated in Fig.\ref{framework}, which has two key components: 1) Importance-aware Parameter Regularization (IPR) for the shared adapter to explain the sensitivity of parameters to previous tasks and adaptively preserve old knowledge in feature space, and 2) the Trainable Semantic Drift Compensation (TSDC) to reduce decision boundary confusion in the unified classifier.
During training, the backbone remains frozen while the shared adapter is updated with IPR, and the unified classifier is refined using TSDC. The IPR component mitigates catastrophic forgetting without adding parameters, maintaining past knowledge in feature space while preserving model flexibility. This involves two steps: 1) calculating each parameter’s importance according to global and local aspects, and 2) applying importance-aware weighted constraints on the shared adapters.
Although IPR adaptively preserve old knowledge in feature space, it can still encounter task boundary confusion in the unified classifier. To address this, TSDC is introduced to train the unified classifier by compensating old prototypes based on the regularized and estimated semantic drift.
\subsection{Adapter-based Tuning Method in Pre-trained Model}
Given a pre-trained Transformer model of $N_{L}$ layers, we learn a set of  adapters to serve as the tunable parameters to adapt the foundation model to the downstream CIL tasks.
In this method, the adapter module $\mathcal{A}^{t}=\{\mathcal{A}_{l}^{t}\}_{l=1}^{N_{L}}$ is utilized with a frozen ViT model serving as the feature extractor. An adapter is a encoder-decoder architecture \cite{chen2022adaptformer}, integrating with a pre-trained transformer network to facilitate transfer learning and enhance downstream task performance. Typically, it consists of a down-sampling MLP layer $\mathbf{W}_{down}\in \mathbb{R}^{d\times d_{h}}$, a ReLU activation function $\bm{\phi}(\cdot)$, and an up-sampling MLP layer $\mathbf{W}_{up} \in \mathbb{R}^{d_{h}\times d}$, where $d$ denotes the input dimension of $\mathbf{W}_{down}$ and output dimension of $\mathbf{W}_{up}$. $d_{h}$ represents the dimension of its hidden units. For adapter, if input is $\mathbf{x}_{i}$, the output is
\begin{equation}\label{eq:important}
    \hspace{2cm}\mathbf{y}_{i}=\bm{\phi} (\mathbf{x}_{i}\cdot\mathbf{W}_{down})\cdot \mathbf{W}_{up}.
\end{equation}
The output of the feature extractor is $\mathbf{f}=\mathcal{F}(\mathbf{x}; \mathcal{A}^{t})$.
\subsection{Importance-aware Parameter Regularization}
During adapter tuning of a pre-trained foundation model, adapters often over-fit to the current downstream tasks. When trained sequentially with dynamically arriving data, these task-shared adapters typically experience significant knowledge forgetting in feature space.
To address this forgetting, existing regularization-based methods often apply uniform, rigid constraints on adapter parameters. While these approaches help maintain model stability, they usually sacrifice too much model plasticity.
To further address these issues,
we propose the Importance-aware Parameter Regularization (IPR) module for the shared adapter during continual adapter tuning. The regularization loss is computed as follow:
\begin{equation}\label{eq:IPR}
\begin{aligned}
\hspace{1cm}\mathcal{L}_{IPR}^{t}=&\sum_{l=1}^{N_{L}}\mathbf{I}_{l,t-1}^{dm}\cdot(\bm{\theta}_{l,t-1}^{dm}-\bm{\theta}_{l,t}^{dm})^{2}\\
    +&\sum_{l=1}^{N_{L}}\mathbf{I}_{l,t-1}^{um}\cdot((\bm{\theta}_{l,t-1}^{um})^{\top}-(\bm{\theta}_{l,t}^{um})^{\top})^{2},
\end{aligned}
\end{equation}
where $\mathbf{I}_{l,t-1}^{dm}\in \mathbb{R}^{d\times d_{h}}$, $\mathbf{I}_{l,t-1}^{um}\in \mathbb{R}^{d\times d_{h}}$ respectively denotes the parameter's importance of dowm-sampling matrix and up-sampling matrix under the $(t-1)$-th task. $N_{L}$ represents the number of adapter modules. The parameters $\bm{\theta}_{l,t-1}^{dm}$ and $\bm{\theta}_{l,t}^{dm}$ correspond to the parameter in the down-sampling matrix of the $l$-th adapter module for the $(t-1)$-th, $t$-th task, respectively. Similarly, $\bm{\theta}_{l,t-1}^{um}$ and $\bm{\theta}_{l,t}^{um}$ represent the parameter in the up-sampling matrix of the $l$-th adapter module for the $(t-1)$-th, $t$- th task. Note that, during model training, the $(t-1)$-th model parameters $\bm{\theta}_{l,t-1}^{dm}$ and $\bm{\theta}_{l,t-1}^{um}$ are fixed, and only current model parameters are trainable.\\
\noindent \textbf{Parameter Importance}: Parameter importance synthesizes two components: 1) \textbf{Global Importance}:
Quantifies the aggregate contribution of all adapter modules by formulating an optimization objective that maximizes the classification score while minimizing variance across different channels of the final output feature.
2) \textbf{Local Importance}: Captures the  specific contribution of individual adapter modules by analyzing the impact of adding perturbations to specific modules on the output.\\
\noindent\textbf{Global importance}. The final output feature, which aggregates responses from all adapter modules, serves as a measure of global importance. By the Central Limit Theorem, features adhering to the same distribution (i.e., belonging to the same class) exhibit channel-wise activations approximating distinct normal distributions under sufficient training samples. Consider a channel $x\sim\mathcal{N}(\mu, \sigma^2)$. For a linear classifier, the decision score $s$ is modeled as:
\begin{equation}\label{eq:linear combination}
   \hspace{3cm} s=w\cdot x + b,
\end{equation}
where $w$ denotes the channel weight vector and $b$ the bias. To optimize classification confidence and stability, we maximize the expected score $E[s]$ while minimizing its variance $V(s)$, introducing a trade-off parameter $\lambda$:
\begin{equation}\label{eq:linear combination}
    \hspace{2.4cm}\max_w{(E[s]-\lambda \cdot V(s))}.
\end{equation}
Substituting $E[s]=w\mu + b$ and $V(s)=w^2\sigma^2$, the objective becomes:
\begin{equation}\label{eq:objective function}
    \hspace{2.2cm}J(w)=w\mu + b - \lambda w^2\sigma ^2.
\end{equation}
Solving $\frac{\partial J}{\partial w}  = \mu - 2\lambda w \sigma^2 = 0$ yields the optimal weight:
\begin{equation}\label{eq: optimal weight}
   \hspace{3cm} w^{*} = \frac{\mu}{2\lambda \sigma^2}.
\end{equation}
Thus, $w^{*}$ is proportional to $\frac{\mu}{\sigma^2}$, indicating that channels with higher mean activations ($\mu$) and lower variance ($\sigma^{2}$) dominate classification decisions. We therefore define $\frac{\mu}{\sigma^2}$ as the global importance of each channel. For the $t$-th task, global importance $\mathbf{G}_{t}\in \mathbb{R}^{1\times d}$ is computed incrementally as:
\begin{equation}\label{eq:channle importance}
\begin{aligned}
    \hspace{1.5cm} \mathbf{G}_{t}=\mathbf{G}_{t-1}+\frac{1}{N_{C}^{t-1}}\sum_{c=1}^{N_{C}^{t-1}}\frac{|\mathbf{f}_{c}|}{\bm{\sigma}_{c}^{2}}, \\
     &\hspace{-4.5cm}\left\{
     \begin{aligned}
         &\mathbf{f}_{c}=\frac{1}{n_{c}}\sum_{i=1}^{n_{c}}\hat{\mathbf{f}}_{i,c}, \\
         & \bm{\sigma}_{c}^{2}=\frac{1}{n_{c}}\sum_{i=1}^{n_{c}}(\hat{\mathbf{f}}_{i,c}-\mathbf{f}_{c})^{\top}(\hat{\mathbf{f}}_{i,c}-\mathbf{f}_{c}), \\
     \end{aligned}
         \right.  
\end{aligned}
\end{equation}
where $\hat{\mathbf{f}}_{i,c}=\mathbf{\mathcal{F}}(\mathbf{x}_{i,c};\mathcal{A}^{t-1})$ is the feature for the sample of class $c$, extracted by the $(t-1)$-th feature extractor $\mathcal{F}$. Here, $\mathbf{f}_{c} \in \mathbb{R}^{1 \times d}$ and $\bm{\sigma}_{c}^{2} \in \mathbb{R}^{1 \times d}$ denote the mean and variance of class $c$, $\mathbf{x}_{i,c}$ is the $i$-th input for the $c$-th class, $n_{c}$ is the sample count for class $c$, and $N_{C}^{t-1}$ is the total classes in the $(t-1)$task . \par
\noindent\textbf{Local importance}. While global importance reflects the aggregate contribution of adapter modules to the final output feature, local importance quantifies the influence of individual adapters embedded at distinct network depths within the pre-trained ViT. To isolate this specific contribution, we analyze perturbations in adapter parameters and their propagation through the network.\\
Consider that a set of adapter modules are combined into a model through concatenation, Under the $l$-th adapter module, the down-sampling matrix $\mathbf{W}^{dm}_{l}\in \mathbb{R}^{d\times d_h}$ and up-sampling matrix $\mathbf{W}^{um}_{l}\in \mathbb{R}^{d_h\times d}$ in the $l$-th layer, with intermediate activation $\mathbf{h}_{l}$, input $\mathbf{x}_l$ and $\mathbf{x}_{l+1}$. A perturbation $\Delta\mathbf{W}^{dm}_{l}$ alters $\mathbf{h}_{l}$, propagating to the final output $\mathbf{x}_L$:
\begin{equation}\label{dowm_importance_1}
    \hspace{1.5cm}\Delta \mathbf{x}_L=\left(\prod_{m=l+1}^{L-1}\frac{\partial \mathbf{x}_{m+1}}{\partial \mathbf{x}_m}\right)\mathbf{W}^{um}_{l}\Delta \mathbf{h}_l.
\end{equation}
where $\Delta \mathbf{h}_l$ is the variation of the intermediate unit $\mathbf{h}_l$. 
Assuming that the activation function is ReLU, and only activated units are considered, $\Delta \mathbf{x}_L$ approximates first-order Taylor expansion 
\begin{equation}\label{dowm_importance_Taylor}
\begin{aligned}
    &\hspace{1cm}\Delta \mathbf{x}_L \approx \frac{\partial \mathcal{F}}{\partial \mathbf{h}_l}\Delta \mathbf{h}_l,\\
    \text{with}  \quad \frac{\partial \mathcal{F}}{\partial \mathbf{h}_l} =& \left[\prod_{m=l+1}^{L-1}\left(\mathbf{I}+\frac{\partial (\mathbf{W}^{um}_{m}\mathbf{h}_{m})}{\partial \mathbf{x}_m}\right)\right]\mathbf{W}^{um}_{l}.
\end{aligned}
\end{equation}
Therefore, $\Delta \mathbf{x}_L$ depends linearly on $\Delta\mathbf{h}_l$, establishing $\mathbf{h}_l$ as a proxy for the local importance of $\mathbf{W}^{dm}_{l}$. For the up-sampling matrix $\mathbf{W}^{um}_{l}$, a perturbation $\Delta\mathbf{W}^{um}_{l}$ directly affects $\mathbf{x}_{L}$:
\begin{equation}\label{up_importance_1}
    \hspace{1.5cm}\Delta \mathbf{x}_L=\left(\prod_{m=l+1}^{L-1}\frac{\partial \mathbf{x}_{m+1}}{\partial \mathbf{x}_m}\right)\Delta \mathbf{W}^{um}_{l}\mathbf{h}_l.
\end{equation}
Similarly, approximating via Taylor expansion yields:
\begin{equation}\label{up_importance_Taylor}
\begin{aligned}
    &\Delta \mathbf{x}_L \approx \frac{\partial \mathcal{G}}{\partial \mathbf{h}_l}\Delta \mathbf{W}^{um}_{l}\mathbf{h}_l,\\
   \hspace{1cm} \text{with}  \quad &\frac{\partial \mathcal{G}}{\partial \mathbf{h}_l} = \prod_{m=l+1}^{L-1}\left(\mathbf{I}+\frac{\partial (\mathbf{W}^{um}_{m}\mathbf{h}_{m})}{\partial \mathbf{x}_m}\right).
\end{aligned}
\end{equation}
Thus, the local importance of $\mathbf{W}^{um}_{l}$ which is governed by the interaction between $\mathbf{h}_l$ and its connection weights is $\mathbf{h}_l\cdot \sum_{k=1}^{d}\mathbf{W}^{um}_{l}[:,k]$.\\
To operationalize this, we compute specific intermediate features for each adapter. For input  $\mathbf{x}_i$  the $l$-th adapter’s intermediate units $\tilde{\mathbf{u}}_{l,t} \in \mathbb{R}^{d_{t} \times d_{h}}$ are derived as
\begin{equation} \label{eq:Intermediate feature}
    \hspace{2cm}\tilde{\mathbf{u}}_{l,t}=\bm{\phi}(\mathbf{x}_{i}\cdot\mathbf{W}_{down,l,t}),
\end{equation}
where $d_{t}$ is the token count. To mitigate information loss from non-[CLS] tokens, we introduce a similarity router that computes a weighted sum of tokens based on cosine similarity to the [CLS] token $\tilde{\mathbf{u}}_{l,t}[0,:]$:
\begin{equation} \label{eq:node feature}
    \hspace{1cm}\mathbf{u}_{l,t}=\sum_{j=0}^{d_{t}}(\cos(\tilde{\mathbf{u}}_{l,t}[j,:],\tilde{\mathbf{u}}_{l,t}[0,:])\cdot \tilde{\mathbf{u}}_{l,t}[j,:]),
\end{equation}
yielding a condensed representation $\mathbf{u}_{l,t} \in \mathbb{R}^{1 \times d_{h}}$. $\cos(\cdot,\cdot)$ denotes the cosine similarity. \\
The local importance of the $l$-th layer ’s down-sampling matrix for task $t$ is then:
\begin{equation} \label{eq:down importance}
   \hspace{1.7cm} \mathbf{L}_{l,t}^{dm}=\mathbf{L}_{l,t-1}^{dm} + \frac{1}{N_t}\sum_{i=1}^{N_t}\mathbf{u}_{l,t,i},
\end{equation}
where $N_{t}$ is the input count for task $t$. For the local importance of up-samping matrix, we first calculate
\begin{equation}\label{eq:up importance_unit}
    \hspace{1.7cm}\hat{\mathbf{u}}_{l,t}=\mathbf{u}_{l,t}\cdot\sum_{k=1}^{d}\left|\mathbf{w}_{l,t}[:,k]\right|,
\end{equation}
where $\mathbf{w}_{l,t}[:,k]$ denotes the weights connecting the weighted hidden unit of the $l$-th adapter module to the $k$-th output unit in the task $t$. Therefore, for the up-sampling matrix of the $l$-th layer under the $t$-th task with $N_t$ inputs, the local importance is
\begin{equation} \label{eq:up importance}
    \hspace{1.7cm}\mathbf{L}_{l,t}^{um}=\mathbf{L}_{l,t-1}^{um} + \frac{1}{N_t}\sum_{i=1}^{N_t}\hat{\mathbf{u}}_{l,t,i},
\end{equation}
Finally, global and local importance are combined to determine parameter importance:
\begin{equation}\label{eq:parameters importance}
    \begin{aligned}
      \hspace{2.5cm}  & \mathbf{I}_{l,t}^{um}= \eta_1\mathbf{G}_{t}^{\top}\mathbf{L}_{l,t}^{um}, \\
        &\mathbf{I}_{l,t}^{dm}=\eta_2\mathbf{G}_{t}^{\top}\mathbf{L}_{l,t}^{dm}, \\
    \end{aligned}
\end{equation}
where $\eta_1$ and $\eta_2$ denote hyper-parameter, $\mathbf{I}_{l,t}^{um}, \mathbf{I}_{l,t}^{dm} \in \mathbb{R}^{d\times d_h}$ govern the up-sampling and down-sampling matrix importance for the $l$-th adapter in task $t$. The algorithm for calculating importance is shown in the Algorithm.\ref{alg::1}.
\vspace{-6mm}
\begin{algorithm}
  \caption{Parameters importance of IPR}  
  \label{alg::1}
    \textbf{Input:\\}
    {The inference data $\{\mathbf{x}_i\}_{i=1}^{N}$;\\
    The prototype of old task $\{\mathbf{f}_{i}\}_{i=0}^{C_{t-1}}$;\\
    The parameters importance $\mathbf{I}_{t-1}^{dm}$, $\mathbf{I}_{t-1}^{um}$; \\
    The pre-trained backbone $\mathcal{F}$, classifier $\mathbf{h}$;\\
    The fine-tuned adapters$ \{\mathcal{A}_{l}^{t}\}_{l=1}^{N_{L}}$;\\}
    \textbf{Output:\\}
    {The parameters importance $I^{dm}_{l,t}$, $I^{um}_{l,t}$. \\}
    \For{$\mathbf{x}_i$ \textbf{in} $\{\mathbf{x}_i\}_{i=1}^N$}{
    Compute the global importance $G_{t}$ as Eq.\ref{eq:channle importance} \\
    Extract intermediate hidden units $\tilde{\mathbf{u}}$ of adapters as Eq.\ref{eq:Intermediate feature}\\
    Obtain the weighted hidden unitse $\mathbf{u}$ as Eq.\ref{eq:node feature}\\
    Compute the local importance of down-sample matrix $L^{dm}_{l,t}$ as Eq.\ref{eq:down importance}\\ 
    Compute the local importance of up-sample matrix $L^{um}_{l,t}$ as Eq.\ref{eq:up importance_unit} and Eq.\ref{eq:up importance}\\
    Obtain the fused parameters importance $I^{dm}_{l,t}$, $I^{um}_{l,t}$ as Eq.\ref{eq:parameters importance}\\ 
    }
\end{algorithm}\textbf{}
\vspace{-6mm}
\subsection{Trainable Semantic Drift Compensation}
Despite the IPR mechanism’s ability to preserve historical knowledge within shared adapters, residual semantic drift across tasks introduces decision boundary ambiguity in the feature space. To mitigate this, we propose a Trainable Semantic Drift Compensation (TSDC) module, which adaptively adjusts old-task prototypes using regularized drift estimates. Compared with prior static methods \cite{yu2020semantic}, TSDC not only dynamically adjusts drift contributions based on the relevance of current-task data but also implements a trainable strategy to enhance robustness against estimation inaccuracies.\\
\noindent \textbf{Semantic Drift Estimation.}
The overall semantic drift is estimated by taking a weighted combination of the feature changes in each current sample. Specifically, we calculate the drift of the current sample using the extracted features from both the $(t-1)$-th and $t$-th feature extractors. The similarity weight is derived from the distance between the $(t-1)$-th extracted feature and the stored class prototypes. By integrating these elements, we obtain an estimated drift for the old prototypes as follows,
\begin{equation}\label{eq:compensation}
\begin{aligned}
\hspace{1.7cm}      \widetilde{\mathbf{\Delta}}_{c}^{t-1\rightarrow{t}}=\frac{\sum\bm{\alpha}_{i}\bm{\delta}_{i}^{t-1\rightarrow{t}}}{\sum\bm{\alpha}_{i}},c\notin \mathcal{C}^{t},\\
     &\hspace{-5cm}\left\{
     \begin{aligned}
         & \bm{\delta}_{i}^{t-1\rightarrow{t}}=\hat{\mathbf{f}}_{i}^{t-1}-\hat{\mathbf{f}}_{i}^{t},\\
         & \bm{\alpha}_{i}=\exp(-\frac{(\hat{\mathbf{f}}_{i}^{t-1}-\mathbf{f}_{c}^{t-1})^{2}}{2\bm{\sigma}_{c}^2}), \\
     \end{aligned}
         \right.  
\end{aligned}
\end{equation}
where $\bm{\delta}_{i}^{t-1\rightarrow{t}}$ represents the semantic drift of current samples. $\bm{\sigma}_{c}$ is the standard deviation of the distribution of $c$-class, $\bm{\alpha}$ denotes the weight for each sample drift.
$\mathcal{C}^{t}$ is the the class set of the $t$-th task, and $\hat{\mathbf{f}}_{i}^{t-1}$, $\hat{\mathbf{f}}_{i}^{t}$ correspond to the $i$-th sample feature of current task extracted by the $(t-1)$-th, $(t)$-th feature extractors.\\
\noindent \textbf{Trainable Semantic Drift.}
Large-magnitude semantic drift introduces noise into compensation estimates, destabilizing prototype updates and degrading decision boundary precision. To mitigate this, we regularize drift estimates toward the zero subspace via a trainable loss term $\mathcal{L}_{TSD}^{t}$, which penalizes deviations from optimal prototype stability:
\begin{equation}\label{eq:Zreg}
    \hspace{2.3cm} \mathcal{L}_{TSD}^{t}=\Vert \widetilde{\mathbf{\Delta}}_{c}^{{t-1}\rightarrow t} \Vert_2^2,
\end{equation}
\noindent \textbf{Old Prototype Compensation.}
Given the estimated semantic drift for all the old classes, the prototypes will be compensated as follows,
\begin{equation}\label{eq: semantic drift}
    \hspace{1.3cm}\mathbf{f}_{c}=\left\{
    \begin{aligned}
        & \mathbf{f}_{c}^{t-1}+\widetilde{\mathbf{\Delta}}_{c}^{t-1\rightarrow{t}}, &\quad c \notin \mathcal{C}^{t}, \\
        & \frac{1}{n_{c}}\sum_{i=1}^{n_{c}}\hat{\mathbf{f}}_{i,c}, &c\in \mathcal{C}^{t}. \\
    \end{aligned}
        \right.
\end{equation}
By employing this compensation update mechanism, we simulate the semantic drift of features in prior tasks within the new feature space, thereby establishing a more precise sampling domain for the subsequent training of a unified classifier.
\begin{algorithm}
  \caption{Training for unified classifier}  
  \label{alg::2}
    \textbf{Input:\\}
    {The inference data $\{\mathbf{x}_i\}_{i=1}^{N}$;\\
    The prototype of old task $\{\mathbf{f}_{i}\}_{i=0}^{C_{t-1}}$;\\
    Sample space $\mathcal{S}$\\
    The pre-trained backbone $\mathcal{F}$, classifier $\{\mathbf{h}_i\}_{i=1}^{t}$;\\
    The fine-tuned adapters$ \{\mathcal{A}_{l}^{t-1}\}_{l=1}^{N_{L}}$, $ \{\mathcal{A}_{l}^{t}\}_{l=1}^{N_{L}}$;\\}
    \textbf{Output:\\}
    {The loss of traing unified classifier $\mathcal{L}_{uc}$. \\}
    \For{$\mathbf{x}_i$ \textbf{in} $\{\mathbf{x}_i\}_{i=1}^N$}{
    Weighted estimate the value of compensation $\widetilde{\mathbf{\Delta}}_{c}^{t-1\rightarrow{t}}$ as Eq.\ref{eq:compensation}\\ 
    Regularize the semantic drift by minimizing the regularization loss $\mathcal{L}_{TSD}^{t}$ as Eq.\ref{eq:Zreg}\\  
    }
    Compensate the old prototype by the regularized drift to get the updated old prototype $\mathbf{f}_c$ as Eq. \ref{eq: semantic drift}\\ 
    Sample old feature $\mathbf{v}_c$ by updated prototype in sample sapce $\mathcal{S}$;\\
    Determine the number of samples for each category $N_{s}$ and obtain the category set $\mathcal{C}$\\
    \For{$\mathbf{v}_{i}$ in $\{\mathbf{v}_{i}\}_{i=0}^{N_{s}\ast |\mathcal{C}|}$}{
    Obtain and minimizing loss of training unified classifier $\mathcal{L}_{uc}$ as Eq.\ref{eq:unified classifier}\\ 
    }
\end{algorithm}\textbf{}

\begin{table*}[t]
\centering
\caption{Experimental results for 10 incremental sessions on four CIL benchmarks.We report the averaged results over 3 random number seeds. The highest results are in bold, and the second highest results are underlined.}
\setlength{\tabcolsep}{2pt} 
\scalebox{0.83}{
\begin{tabular}{cccccccccc}
\toprule
\multirow{2}*{Method} &  \multirow{2}*{Venue} &\multicolumn{2}{c}{Split-ImageNetR} &\multicolumn{2}{c}{Split-ImageNetA} &\multicolumn{2}{c}{CUB200} &\multicolumn{2}{c}{CIFAR100}\\
&{}
&$A_{Last} \uparrow $
&$A_{Avg} \uparrow $
&$A_{Last} \uparrow $
&$A_{Avg} \uparrow $ 
&$A_{Last} \uparrow $
&$A_{avg} \uparrow $ 
&$A_{Last} \uparrow $
&$A_{avg} \uparrow $\\
\hline
L2P \cite{wang2022learning}  & CVPR’22     &$72.34_{\pm 0.17}$& $77.36_{\pm 0.64}$ &$44.04_{\pm 0.93}$&$51.24_{\pm 2.26}$ &$67.02_{\pm 1.90}$ &$79.62_{\pm 1.60}$&	   $84.06_{\pm 0.88}$ &$88.26_{\pm 1.34}$\\
DualPrompt \cite{wang2022dualprompt}  & ECCV’22   & $69.10_{\pm 0.62}$ &  $74.28_{\pm 0.66}$ &$53.19_{\pm 0.74}$ &  $64.59_{\pm 0.08}$ &  $68.48_{\pm 0.47}$ &$80.59_{\pm 1.50}$ &  $86.93_{\pm 0.24}$ &$91.13_{\pm 0.32}$  \\
ADA \cite{ermis2022memory}   & NeurIPS’
22    &$73.76_{\pm 0.27}$& $79.57_{\pm 0.84}$ &$50.16_{\pm 0.20}$&$59.43_{\pm 2.20}$ &$76.13_{\pm 0.94}$ &$85.74_{\pm 0.26}$&	   $88.25_{\pm 0.26}$ &$91.85_{\pm 1.32}$\\
CODAPrompt \cite{CODAPromptCOntinual2023Smith}  & CVPR’23   & $73.31_{\pm 0.50}$ &  $78.47_{\pm 0.53}$ &$52.08_{\pm 0.12}$ &  $63.92_{\pm 0.12}$ &  $77.23_{\pm 1.12}$ &$81.90_{\pm 0.85}$  &  $83.21_{\pm 3.39}$ &$87.71_{\pm 3.17}$  \\

SLCA \cite{zhang2023slca}    & ICCV’23     & $79.35_{\pm 0.28}$	&$83.29_{\pm 0.46}$& $61.05_{\pm 0.63}$&	$68.88_{\pm 2.31}$ &$84.68_{\pm 0.09}$&	$90.77_{\pm 0.79}$ &  $ 91.26_{\pm 0.37}$  & $94.29_{\pm 0.92}$ \\
LAE \cite{gao2023unified} & ICCV’23 
  & $72.29_{\pm 0.14}$	&$77.99_{\pm 0.46}$& $47.18_{\pm 1.17}$	&$58.15_{\pm 0.73}$ & $80.97_{\pm 0.51}$	&$87.22_{\pm 1.21}$&$85.25_{\pm 0.43}$ &$89.80_{\pm 1.20}$ \\
Adam-adapter \cite{zhou2024revisiting} & IJCV’24 & $65.79_{\pm 0.98}$	&$72.42_{\pm 1.41}$& $48.81_{\pm 0.08}$&$58.84_{\pm 1.37}$ &$85.84_{\pm 0.08}$&	$91.33_{\pm 0.49}$   &$87.29_{\pm 0.27}$&	$91.21_{\pm 1.33}$   \\

DualP-PGP \cite{PromptGradient2024Qiao}& ICLR’24  & $76.11_{\pm 0.21}$ &  $-$ &$-$ &  $-$ &  $-$ &$-$  &  $86.92_{\pm 0.05}$ &$-$ \\
OS-Prompt++ \cite{kim2024one} & ECCV’24  &$75.67_{\pm 0.40}$ &$-$ &$-$ &  $-$ &  $-$ &$-$ &  $86.68_{\pm 0.67}$ & $-$   \\
InfLoRA \cite{liang2024inflora}   & CVPR’24 & $75.65_{\pm 0.14}$ &  $80.82_{\pm 0.24}$ &$-$ &  $-$ &  $-$ &$-$  &  $86.51_{\pm 0.73}$ &$91.70_{\pm 0.32}$  \\
CPrompt \cite{ConsistentPrompting2024Gao}& CVPR’24  &$77.14_{\pm 0.11}$ &  $82.92_{\pm 0.70}$ &$-$ &  $-$ &  $-$ &$-$ &  $87.82_{\pm 0.21}$ & $92.53_{\pm 0.23}$  \\
SSIAT \cite{tan2024semantically}  & CVPR’24    &$79.38_{\pm 0.59}$ &  $\underline{83.63}_{\pm 0.43}$ &$\underline{62.43}_{\pm 1.63}$ &  $\underline{70.83}_{\pm 1.63}$ &  $\underline{88.75}_{\pm 0.38}$ &$\underline{93.00}_{\pm 0.90}$ &  $\underline{91.35}_{\pm 0.26}$ & $\underline{94.35}_{\pm 0.60}$    \\
VPT-NSP \cite{lu2024visual} & NeurIPS’24  &$77.95_{\pm 0.22}$ &  $83.44_{\pm 0.40}$ &$53.83_{\pm 0.37}$ &  $63.93_{\pm 1.08}$ &  $83.08_{\pm 0.75}$ &$90.32_{\pm 0.62}$ &  $90.19_{\pm 0.43}$ & $94.12_{\pm 1.05}$ \\
\hline
\textbf{EKPC (Ours)}  & $\mathbf{-}$  & $\mathbf{80.60_{\pm 0.08}}$ &  $\mathbf{84.92_{\pm 0.48}}$ &$\mathbf{64.56_{\pm 0.68}}$ &  $\mathbf{72.09_{\pm 1.57}}$ &  $\mathbf{90.33_{\pm 0.59}}$ &$\mathbf{93.94_{\pm 0.63}}$ &  $\mathbf{92.14_{\pm 0.03}}$ & $\mathbf{94.65_{\pm 0.85}}$   \\ 
\bottomrule
\end{tabular}}
\label{fourdataset}
\end{table*}
\subsection{Loss Function and Optimization}
The overall loss function in our framework contains: 1) the cosine loss function acts as the overall classification loss; 2) the proposed importance-aware parameter regularization; 3) the proposed semantic drift regularization loss. Specifically, the cosine loss function \cite{peng2022few, wang2018cosface} can be written as follows:
\begin{equation}
\begin{aligned}
   \mathcal{L}_{cos}^{t}=-\frac{1}{N^t}\sum_{j=1}^{N^t} \log{\frac{{z_j}}{{z_j}+\sum_{i\ne j }\exp(s\cos{(\bm{\theta}_{i})})}},\\
   \text{where} \quad {z_j}=\exp(s(\cos{(\bm{\theta}_{j})}-m)), 
\end{aligned}
\label{classifier loss}
\end{equation}
where $\cos{(\bm{\theta}_{j})}= \frac{\mathbf{w}_{j} \cdot \mathbf{\hat{f}}}{\Vert \mathbf{w}_{j} \Vert \cdot \Vert \mathbf{\hat{f}} \Vert } $, with $\mathbf{w}_{j}$  denoting the weights of the $j$-th class classifier, $N^t$ representing the number of training samples for current task, and $s$ and $m$ indicating the scale and margin factors. \par
Finally, the overall loss function can be written as follows,
\begin{equation}\label{eq:important}
\begin{aligned}
\hspace{1.4cm}\mathcal{L}^{t}=&\mathcal{L}_{cos}^{t}+w_{1}\mathcal{L}_{IPR}^{t}+w_{2}\mathcal{L}_{TSD}^{t},   
\end{aligned}
\end{equation}
where $w_{1}$ and $w_{2}$ are hyper-parameters to balance the three terms. \\ 
\noindent \textbf{Unified Classifier Training}.
After training the model to classify incremental tasks, we freeze the feature extractor and concatenate all classifiers. Then, we retrain an unified classifier. Following previous works~\cite{zhang2023slca,tang2023prompt,zhu2021prototype,tan2024semantically}, we model each class as a Gaussian distribution and sample features from it. The covariance $\mathbf{\Sigma }_{c}$ and the averaged prototypes $\mathbf{f}_{c}$ for previous tasks are updated through Eq.\ref{eq: semantic drift} to define the sample space $\mathcal{S}$. For each class $c$, a sample feature $\mathbf{v}_{i}$ is drawn from the distribution $\mathcal{N}(\mathbf{f}_{c}, \bm{\Sigma }_{c})$ within $\mathcal{S}$. The loss function for training the unified classifier is:
\begin{equation}\label{eq:unified classifier} 
\hspace{1.2cm}    \mathcal{L}_{uc}=-\sum_{i=1}^{N_{s}\cdot |\mathcal{C}|}\log\frac{\exp(\mathbf{h}(\mathbf{v}_{i})[y_i])}{{\textstyle \sum_{k\in \mathcal{C}}}\exp(\mathbf{h}(\mathbf{v}_{i})[k])},
\end{equation}
where $\mathbf{h}$ denotes the unified classifier, and $y_i$ is the class label of input $\mathbf{v}_{i}$. $\mathcal{C}$ is the set of the learned classes, and $N_{s}$ represents the number of sample features per class. The algorithm for training a unified classifier is shown in the Algorithm.\ref{alg::2}.
\section{Experiments}
\begin{table*}[t]
\centering
\caption{Experimental results for 20 incremental sessions on two CIL benchmarks. We report the averaged results
over 3 random number seeds. The highest results are in bold, and the second highest results are underlined.}
\setlength{\tabcolsep}{10pt} 
\scalebox{1.1}{
\begin{tabular}{cccccc}
\toprule
 \multirow{2}*{Method} &\multicolumn{2}{c}{Split-ImageNetR} &\multicolumn{2}{c}{Split-ImageNetA}  \\
 & $A_{Last}\uparrow$& $A_{Avg}\uparrow$&$A_{Last}\uparrow$ &$A_{Avg}\uparrow$ \\
\hline
L2P \cite{wang2022learning}   & $69.64_{\pm 0.42}$	& $75.28_{\pm 0.57}$ & $40.48_{\pm 1.78}$ & $49.62_{\pm 1.46}$ &\\
DualPrompt \cite{wang2022dualprompt} & $66.61_{\pm 0.58}$ & $72.45_{\pm 0.37}$ & $42.28_{\pm 1.94}$ & $53.39_{\pm 1.64}$ & \\
CODAPrompt \cite{CODAPromptCOntinual2023Smith} & $69.96_{\pm 0.50}$ & $75.34_{\pm 0.85}$ & $44.62_{\pm 1.92}$ & $54.86_{\pm 0.50}$ & \\
SLCA \cite{zhang2023slca}   & $74.63_{\pm 1.55}$& $79.92_{\pm 1.29} $ &  $36.69_{\pm 21.31}$ & $56.35_ {\pm 7.09}$         \\
LAE \cite{gao2023unified}     & $69.86_{\pm0.43}$  & $77.38_{\pm 0.61}$ & $39.52_{\pm 0.78}$ & $ 51.75_{\pm 2.15} $\\
Adam-adapter\cite{zhou2024revisiting}    &$57.42_{\pm 0.84}$  & $64.75_{\pm 0.79}$  &  $48.65_{\pm 0.12}$ & $59.55_{\pm 1.07} $ \\
InfLoRA \cite{liang2024inflora} & $71.01_{\pm 0.45}$ & $77.28_{\pm 0.45}$ & $44.62_{\pm 1.92}$ & $54.86_{\pm 0.50}$ & \\ 
CPrompt \cite{ConsistentPrompting2024Gao} & $74.79_{\pm 0.28}$ & $81.46_{\pm 0.93}$ & $-$ & $-$ &\\
OS-Prompt++ \cite{kim2024one}  & $73.77_{\pm 0.19}$ & $-$ & $-$ & $-$ &\\
SSIAT \cite{tan2024semantically} & $75.67_{\pm 0.24}$ & $\underline{82.30}_{\pm 0.36}$ & $\underline{59.16}_{\pm 1.03}$ & $\underline{68.45}_{\pm 1.92}$ & \\ 
VPT-NSP \cite{lu2024visual}  & $\underline{75.69}_{\pm 0.61}$ & $81.87_{\pm 0.59}$ & $49.81_{\pm 1.29}$ & $61.41_{\pm 1.84}$ & \\
\hline
\textbf{EKPC (Ours)} & $\mathbf{79.42_{\pm 0.43}}$ & $\mathbf{84.24_{\pm 0.59}}$ & $\mathbf{62.02_{\pm 1.45}}$ & $\mathbf{70.38_{\pm 1.83}}$ \\ 
\bottomrule
\end{tabular}}
\label{20s}
\end{table*}

\subsection{Experimental Details}
\noindent \textbf{Dataset.}
Our proposed method is evaluated on five widely-used CIL benchmark datasets: ImageNetR  \cite{hendrycks2021many}, ImageNetA \cite{hendrycks2021natural}, CUB-200 \cite{wah2011caltech}, CIFAR100 \cite{krizhevsky2009learning}, and Domainet\cite{peng2019moment}. ImageNetR includes 30,000 images from 200 classes, sharing class labels with ImageNet-21K but representing different domains. ImageNetA comprises 7,500 images containing challenging samples with significant class imbalance. CUB-200 has 11,788 images of 200 bird species, tailored for fine-grained classification focusing on subtle visual distinctions. CIFAR100 is another commonly used dataset in continual learning, consisting of 60,000 32×32 images across 100 classes. Domainet is a cross-domain dataset comprising images from 345 classes spanning six diverse domains. Following established practices \cite{lu2024visual}, we select the 200 categories with the most images for training and evaluation.\\
\noindent \textbf{Evaluation Metrics. }
In continual learning, the performance of classification models is typically evaluated using two key metrics: 1) Last Accuracy ($A_{Last}$), which measures the accuracy across all classes after completing the final task, and 2) Average Accuracy ($A_{Avg}$), which calculates the average accuracy achieved across all tasks during the learning process.\\
\noindent \textbf{Implementation Details.}
The experiments are conducted on a single NVIDIA RTX 3090 GPU. The model is trained using the SGD optimizer, with an initial learning rate of 0.01, a weight decay of 0.0005, and a mini-batch size of 48. The training process consist of 30 epochs during the first session, followed by 15 epochs in each subsequent session, and the unified classifier is further trained for an additional 5 epochs. Consistent with previous studies, we adopt a ViT-B/16 model pre-trained on ImageNet-21K as the backbone. The scale parameters $s$ and $m$ in Eq.\ref{classifier loss} for the cosine classifier are set to 20.0 and 0.01, respectively. For Eq.\ref{eq:down importance} and Eq.\ref{eq:up importance}, the scale parameters $\eta_1$ and $\eta_2$ are set to 100.0 and 1.0, respectively. The scale $w_1$ and $w_2$ in Eq.\ref{eq:important} are both set to 1.0. In the experiments, we report the mean and standard deviation of the results across three runs for each dataset.

\begin{figure*}[htbp!]
\centering
\includegraphics[width=1\linewidth]{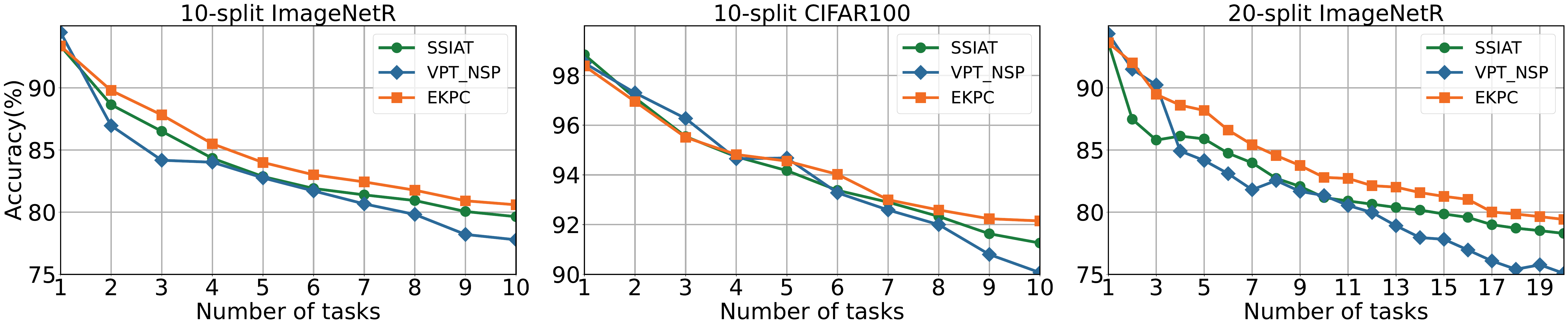}
\caption{The performance of each session on ImageNetR and CIFAR100}
\label{Fig: various_methods}
\end{figure*}

\begin{table}[t]
\centering
\caption{Experimental results for 10 incremental sessions on Split-DmainNet dataset. We report the averaged results over 3 random number seeds. The highest results are in bold, and the second highest results are underlined.}
\setlength{\tabcolsep}{2pt} 
\scalebox{0.8}{
\begin{tabular}{ccccccccccccc}
\toprule
 \multirow{2}*{Method} &  \multirow{2}*{Venue} &  \multicolumn{2}{c}{Split-DomainNet}\\
{}&{}&$A_{Last} \uparrow $&$A_{Avg} \uparrow$\\
\hline
L2P \cite{wang2022learning}     &CVPR’22   & $81.17_{\pm 0.83}$	& $87.43_{\pm 0.95}$\\
DualPrompt \cite{wang2022dualprompt}  &ECCV’22  & $81.70_{\pm 0.78}$	& $87.80_{\pm 0.99}$\\
CODAPrompt \cite{CODAPromptCOntinual2023Smith}  &CVPR’23  & $80.04_{\pm 0.79}$	&$86.27_{\pm 0.82}$ \\
DualP-PGP \cite{PromptGradient2024Qiao}  &ICLR’24 & $80.41_{\pm 0.25}$	& $-$\\
InfLoRA \cite{liang2024inflora} &CVPR’24   & $74.53_{\pm 0.23}$	& $79.57_{\pm 0.57}$\\
CPrompt \cite{ConsistentPrompting2024Gao} &CVPR’24 & $82.97_{\pm 0.34}$	& $88.54_{\pm 0.41}$ \\
SSIAT \cite{tan2024semantically}  &CVPR’24  & $\underline{85.11}_{\pm 0.56}$	& $\underline{89.80}_{\pm 0.34}$\\
VPT-NSP \cite{lu2024visual}  &NeurIPS’24 & $83.05_{\pm 1.06}$	& $87.83_{\pm 0.92}$\\
\hline
\textbf{EKPC (Ours)} &--  & $\mathbf{87.22_{\pm 0.06}}$	& $\mathbf{91.14_{\pm 0.08}}$  \\ 
\bottomrule
\end{tabular}}
\vspace{-4.0mm}
\label{domainnet}
\end{table}

\begin{table*}[t]
\centering
\caption{The Last-acc for long-term class incremental learning setting (50 \& 100 tasks) on two CIL benchmarks. The highest results are in bold and the second highest results are underlined.}
\label{tab: long-term}
\setlength{\tabcolsep}{15pt} 
\scalebox{1}{
\begin{tabular}{ccccc}
\toprule
\multirow{2}*{Method} & 
\multicolumn{2}{c}{Split-ImageNetR} &
\multicolumn{2}{c}{Split-Domainet}\\

& 50s & 100s & 50s & 100s\\
\cline{2-5}
\hline
L2P\cite{wang2022learning} & 51.38 & 41.51 & 63.13 & 54.83 \\
OVOR-Deep\cite{OVOROnePrompt2024Huang} & 63.25 & 43.02 & 68.29 & 52.09\\
ConvPrompt\cite{roy2024convolutional} & 64.61 & 44.32 & 71.76 & 56.21\\
InfLoRA\cite{liang2024inflora} & 62.81 & 42.23 & \underline{71.87} &48.06 \\
Cprompt\cite{ConsistentPrompting2024Gao} & \underline{70.75} & 59.90 & 70.74 & \underline{57.60}\\
EASE \cite{ExpandableSubspace2024Zhou} & 70.27 & 51.56 & 65.34 & 37.56\\
VPT-NSP \cite{lu2024visual} & 69.48 & \underline{62.23} & 71.28 & 57.35 \\
\hline
\textbf{EKPC (Ours)} & \textbf{75.88} & \textbf{70.35} & \textbf{84.59} & \textbf{81.24}\\
\bottomrule
\end{tabular}}

\label{long}
\end{table*}

\begin{table*}[t]
\centering
\caption{Ablation Studies for 10 incremental sessions of each component in our proposed method on three CIL benchmarks. We report the averaged results over 3 random number seeds. The highest results are in bold.}\label{tab: ablation-study}
\setlength{\tabcolsep}{7pt} 
\scalebox{1}{
\begin{tabular}{c@{\hspace{5pt}}c@{\hspace{5pt}}c@{\hspace{5pt}}c@{\hspace{5pt}}c@{\hspace{5pt}}c@{\hspace{10pt}}cccccccc}
\toprule
\multirow{2}*{Idx} & 
\multirow{2}*{$\mathcal{A}^t$} &
\multirow{2}*{IPR} & 
\multirow{2}*{TSDC} &
\multicolumn{2}{c}{Split-DomainNet} &
\multicolumn{2}{c}{Split-ImageNetA} & 
\multicolumn{2}{c}{Split-ImageNetR}\\

& {} & {} & {} 
& $A_{Last}$ $\uparrow$ &$A_{Avg}$ $\uparrow$
& $A_{Last}$ $\uparrow$ &$A_{Avg}$ $\uparrow$
& $A_{Last}$ $\uparrow$ &$A_{Avg}$ $\uparrow$\\
\hline
1& \checkmark & -- & -- &82.75\tsb{\(\pm\)0.50} & 87.50\tsb{\(\pm\)0.66} & 57.08\tsb{\(\pm\)1.65} & 66.64\tsb{\(\pm\)2.11} & 77.71\tsb{\(\pm\)0.44} & 82.74\tsb{\(\pm\)0.75}\\
2& \checkmark & \checkmark & -- &83.53\tsb{\(\pm\)0.10} & 88.78\tsb{\(\pm\)0.66} & 59.14\tsb{\(\pm\)1.87} & 67.98\tsb{\(\pm\)2.40} & 78.75\tsb{\(\pm\)0.38} & 83.92\tsb{\(\pm\)0.84}\\
3& \checkmark & -- & \checkmark &86.24\tsb{\(\pm\)0.42} & 90.82\tsb{\(\pm\)0.23} & 64.10\tsb{\(\pm\)1.27} & 71.62\tsb{\(\pm\)0.48} & 80.42\tsb{\(\pm\)0.15} & 84.77\tsb{\(\pm\)0.55}\\
4& \checkmark & \checkmark & \checkmark &
\textbf{87.22}\tsb{\(\pm\)0.06} & \textbf{91.14}\tsb{\(\pm\)0.08} & \textbf{64.56}\tsb{\(\pm\)0.68} &  \textbf{72.09}\tsb{\(\pm\)1.57} &
\textbf{80.60}\tsb{\(\pm\)0.08} & \textbf{84.92}\tsb{\(\pm\)0.48}\\
\bottomrule
\end{tabular}}
\end{table*}

\begin{table*}[t]
\centering
\caption{Experimental results for 10-tasks of Global (GI) and Local (LI) Importance on
Split-ImageNetA and Split-ImageNetR benchmarks. We report the averaged results over 3 random number seeds. The highest results are in bold.}
\label{tab:GI_and_LI}
\setlength{\tabcolsep}{10pt} 
\scalebox{1}{
\begin{tabular}{c@{\hspace{10pt}}c@{\hspace{10pt}}c@{\hspace{10pt}}c@{\hspace{20pt}}cccccccc}
\toprule
\multirow{2}*{Idx} & 
\multirow{2}*{$\mathcal{A}^t$} &
\multirow{2}*{GI} & 
\multirow{2}*{LI} &
\multicolumn{2}{c}{Split-ImageNetA} & 
\multicolumn{2}{c}{Split-ImageNetR}\\
& {} & {} & {}
& $A_{Last}$ $\uparrow$ &$A_{Avg}$ $\uparrow$
& $A_{Last}$ $\uparrow$ &$A_{Avg}$ $\uparrow$\\
\hline
1& \checkmark & -- & -- & 57.08\tsb{\(\pm\)1.65} & 66.64\tsb{\(\pm\)2.11} & 77.71\tsb{\(\pm\)0.44} & 82.74\tsb{\(\pm\)0.75}\\
2& \checkmark & \checkmark & -- & 58.13\tsb{\(\pm\)2.04} & 67.52\tsb{\(\pm\)2.23} & 77.94\tsb{\(\pm\)0.76} & 83.39\tsb{\(\pm\)1.02} \\
3& \checkmark & -- & \checkmark& 58.02\tsb{\(\pm\)2.06} & 67.47\tsb{\(\pm\)2.05} & 78.36\tsb{\(\pm\)0.31} & 83.76\tsb{\(\pm\)0.80} \\
4& \checkmark & \checkmark  & \checkmark
& \textbf{59.14}\tsb{\(\pm\)1.87} & \textbf{67.98}\tsb{\(\pm\)2.40} & \textbf{78.75}\tsb{\(\pm\)0.38} & \textbf{83.92}\tsb{\(\pm\)0.84}\\
\bottomrule
\end{tabular}}
\end{table*}

\subsection{Comparison With State-of-the-art Methods}
The proposed method consistently achieves superior performance compared to state-of-the-art (SOTA) approaches across various benchmarks, as shown in Table.\ref{fourdataset}, Table.\ref{domainnet} and Table.\ref{20s}. We evaluate our method against prompt-based methods \cite{wang2022learning, wang2022dualprompt, CODAPromptCOntinual2023Smith, PromptGradient2024Qiao, ConsistentPrompting2024Gao, kim2024one, lu2024visual}, adapter(or LoRA)-based methods \cite{ermis2022memory, zhou2024revisiting, liang2024inflora, tan2024semantically}, and other PEFT methods \cite{gao2023unified, zhang2023slca}. Prompt-based methods maintain a set of task-specific prompts and select the most appropriate prompt for each class. Adapter-based methods mitigate catastrophic forgetting in continual learning by integrating either expandable or shared adapters. Table.\ref{fourdataset} and Table.\ref{domainnet} present experimental results on five CIL benchmarks across 10 incremental sessions, while Table.\ref{20s} reports the performance across 20 incremental sessions. Consistent with prior findings \cite{tan2024semantically, zhou2024revisiting}, adapter-based methods generally outperform prompt-based and other PEFT methods. Notably, under the same tuning strategy, the proposed method further surpasses adapter-based approaches, highlighting its superior effectiveness in mitigating catastrophic forgetting and enhancing model adaptability. \\
To further evaluate the effectiveness and robustness of the proposed method in long-term continual learning, we conduct experiments with 50 and 100 tasks on two benchmarks, reporting the Last Accuracy $A_{Last}$. We compare the proposed method with seven existing methods, as detailed in Table.\ref{long}, including prompt-based methods \cite{wang2022learning, roy2024convolutional, ConsistentPrompting2024Gao, lu2024visual, OVOROnePrompt2024Huang}, adapter(or LoRA)-based methods \cite{liang2024inflora}, and expandable subspace ensemble based method \cite{ExpandableSubspace2024Zhou}. The results show that the proposed method outperforms all comparison methods, which demonstrates the effectiveness of the proposed importance-aware regularization strategy in maintaining model stability while promoting adaptive plasticity, thereby effectively balancing the trade-off between stability and plasticity in incremental learning.\\
To further analyze the model's performance across individual tasks, we conduct a detailed per-task analysis in comparison with current state-of-the-art methods. As illustrated in Fig \ref{Fig: various_methods}, comparisons are performed on the ImageNetR and CIFAR100 datasets under a 10-task experimental setting, and additionally on the ImageNetR dataset under a 20-task configuration. The results demonstrate that the proposed method consistently achieves higher accuracy than existing methods across nearly all tasks, with its advantages becoming increasingly pronounced as the number of tasks grows. This superior performance suggests that our approach effectively preserves the knowledge of previously learned tasks while maintaining the plasticity for learning new tasks.\par

\begin{table*}[t]
\centering
\caption{Experimental results for 10-tasks of our IPR method and other parameter regularization methods on three CIL benchmarks. We report the averaged results over 3 random number seeds. The highest results are in bold. }
\label{tab: PR vs Fisher vs ABS vs IPR}
\setlength{\tabcolsep}{7.5pt} 
\scalebox{1}{
\begin{tabular}{c@{\hspace{5pt}}c@{\hspace{5pt}}c@{\hspace{5pt}}c@{\hspace{5pt}}c@{\hspace{5pt}}c@{\hspace{10pt}}ccccccc}
\toprule
\multirow{2}*{Idx} & 
\multirow{2}*{$\mathcal{A}^t$} &
\multirow{2}*{PR} & 
\multirow{2}*{APR} &
\multirow{2}*{FPR} &
\multirow{2}*{IPR} &
\multicolumn{2}{c}{Split-DomainNet} &
\multicolumn{2}{c}{Split-ImageNetA} & 
\multicolumn{2}{c}{Split-ImageNetR}\\
& {} & {} & {} & {} & {}
& $A_{Last}$ $\uparrow$ &$A_{Avg}$ $\uparrow$
& $A_{Last}$ $\uparrow$ &$A_{Avg}$ $\uparrow$
& $A_{Last}$ $\uparrow$ &$A_{Avg}$ $\uparrow$\\
\hline
1& \checkmark & -- & -- & -- & -- &82.75\tsb{\(\pm\)0.50} & 87.50\tsb{\(\pm\)0.66} & 57.08\tsb{\(\pm\)1.65} & 66.64\tsb{\(\pm\)2.11} & 77.71\tsb{\(\pm\)0.44} & 82.74\tsb{\(\pm\)0.75}\\
2& \checkmark & \checkmark & -- & -- & -- &81.19\tsb{\(\pm\)0.43} & 87.08\tsb{\(\pm\)1.23} & 58.22\tsb{\(\pm\)1.21} & 67.39\tsb{\(\pm\)2.36} & 74.42\tsb{\(\pm\)0.50} & 80.34\tsb{\(\pm\)1.06} \\
3& \checkmark & -- & \checkmark& -- &-- & 83.26\tsb{\(\pm\)0.23} & 88.51\tsb{\(\pm\)0.52} & 58.11\tsb{\(\pm\)1.36} & 67.70\tsb{\(\pm\)2.07} & 78.05\tsb{\(\pm\)0.52} & 83.43\tsb{\(\pm\)0.97} \\
4& \checkmark & -- &  --  & \checkmark&  -- & 82.86\tsb{\(\pm\)0.32} & 88.25\tsb{\(\pm\)0.57} & 57.74\tsb{\(\pm\)1.85} & 67.43\tsb{\(\pm\)2.13} & 77.69\tsb{\(\pm\)0.51} & 83.28\tsb{\(\pm\)0.98} \\
5& \checkmark & -- & -- & -- & \checkmark 
& \textbf{83.53}\tsb{\(\pm\)0.10} & \textbf{88.78}\tsb{\(\pm\)0.66} & \textbf{59.14}\tsb{\(\pm\)1.87} & \textbf{67.98}\tsb{\(\pm\)2.40} & \textbf{78.75}\tsb{\(\pm\)0.38} & \textbf{83.92}\tsb{\(\pm\)0.84}\\
\bottomrule
\end{tabular}}
\end{table*}

\begin{table*}
\centering
\caption{Performance comparison of training unified classifier methods: Without-Semantic Drift (Wo-SD), Static semantic drift (Static), and Trainable Semantic Drift 
Compensation (TSDC) approaches. The lowest value of drift and the highest accuracy are in bold.}
\setlength{\tabcolsep}{1pt} 
\scalebox{1.1}{
\begin{tabular}{c|ccc@{\hspace{10pt}}ccc@{\hspace{10pt}}ccc}
\toprule
 \multirow{2}*{Method} &\multicolumn{3}{c}{DomainNet} &\multicolumn{3}{c}{ImageNetA} &\multicolumn{3}{c}{ImageNetR} \\
 & $SDV\downarrow$ & $A_{Last}\uparrow$ & $A_{Avg}\uparrow$ & $SDV\downarrow$ & $A_{Last}\uparrow$ & $A_{Avg}\uparrow$ &  $SDV\downarrow$ & $A_{Last}\uparrow$ & $A_{Avg}\uparrow$\\
\hline
\multirow{1}*{Wo-SD} & -- &83.34\tsb{\(\pm\)0.84} & 88.70\tsb{\(\pm\)0.49} & -- & 59.82\tsb{\(\pm\)1.95} & 68.78\tsb{\(\pm\)1.78} & -- & 78.23\tsb{\(\pm\)0.39} & 83.02\tsb{\(\pm\)0.53}\\
\multirow{1}*{Static} & 0.346 & 85.11\tsb{\(\pm\)0.56} & 89.80\tsb{\(\pm\)0.34}  &0.299 & 62.43\tsb{\(\pm\)1.63} & 70.83\tsb{\(\pm\)1.63} & 0.266 & 79.38\tsb{\(\pm\)0.59}& 83.63\tsb{\(\pm\)0.43} \\
\multirow{1}*{TSDC (ours)} & \textbf{0.010} & \textbf{86.24\tsb{\(\pm\)0.42}} & \textbf{90.82\tsb{\(\pm\)0.23}} & \textbf{0.025}  & \textbf{64.10\tsb{\(\pm\)1.27}} & \textbf{71.62\tsb{\(\pm\)0.48}} & \textbf{0.013}  & \textbf{80.42\tsb{\(\pm\)0.15}}& \textbf{84.77\tsb{\(\pm\)0.55}} \\
\bottomrule
\end{tabular}}
\label{tab: Diff_SD}
\end{table*}


\subsection{Ablation Study}
\begin{figure}[htbp!]
\centering
\vspace{-5mm}
\includegraphics[width=\linewidth]{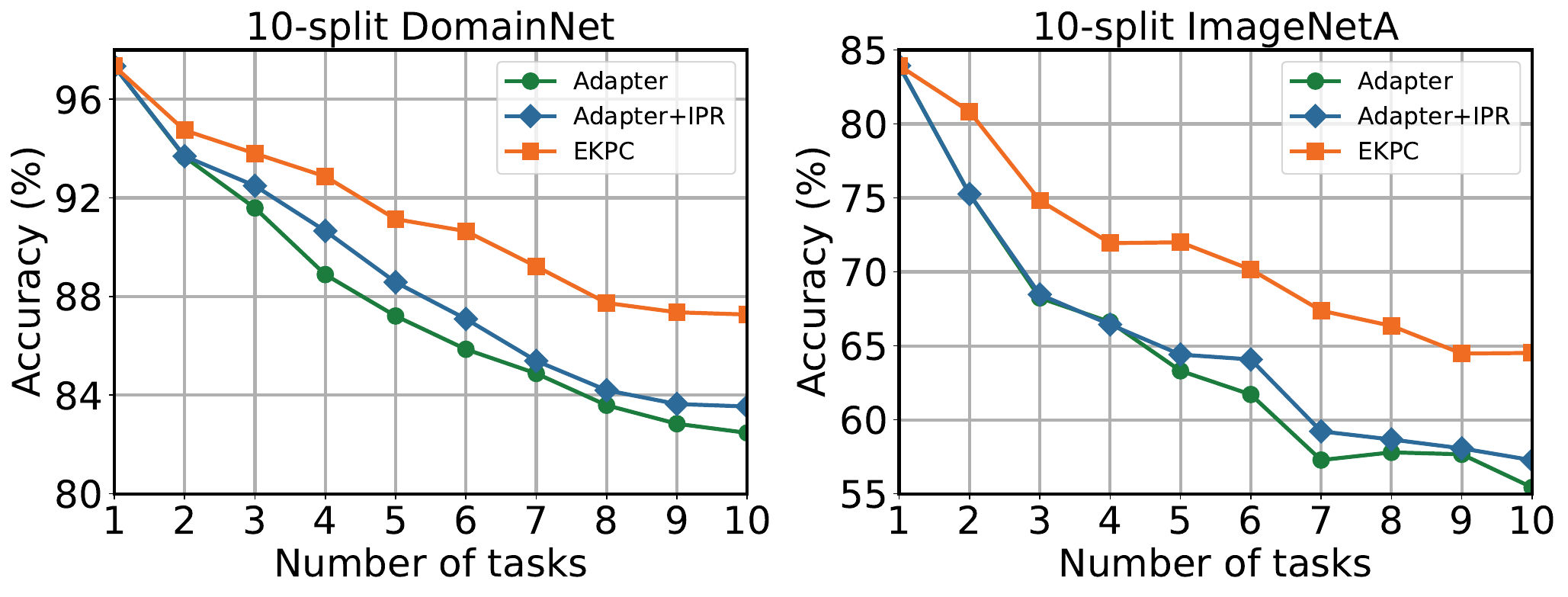}
\caption{The task-by-task accuracy changing curves of our baseline(Adapter), Adapter+IPR, Adapter+IPR+TSDC(EKPC).}
\label{Ablation}
\vspace{-6.0mm}
\end{figure}

\noindent\textbf{The effectiveness of IPR and TSDC.}
Our approach introduces two key innovations: 1) Importance-aware Parameter Regularization (IPR) and 2) Trainable Semantic Drift Compensation (TSDC). To evaluate the effectiveness of these proposed components, we conduct comprehensive ablation study across 10-tasks on three well-established CIL benchmarks, as shown in Table.\ref{tab: ablation-study}. The shared adapter-based method ($\mathcal{A}^t$) serves as the baseline (Idx1). In Idx2, the IPR module is integrated into the baseline model. Experimental results show consistent improvements in both $A_{Last}$ and $A_{Avg}$ across all three datasets, demonstrating the effectiveness of parameter importance regularization in preserving knowledge and mitigating catastrophic forgetting. In Idx3, the TSDC module is incorporated into the baseline. The results confirm that the inclusion of the TSDC module, which effectively mitigates decision boundary confusion across tasks, significantly enhances the performance of the baseline method. Idx4 combines the IPR and TSDC modules, and their collaboration achieves the best performance. Fig \ref{Ablation} visualizes the performance gains from module integration across different sessions on two benchmarks. Notably, as the number of tasks increases, the effectiveness of our approach becomes more pronounced, underscoring its advantages in long-term learning scenarios.\\

\noindent \textbf{The effectiveness of Global and Local Importance.}
The IPR module comprises two key components: Global Importance (GI) and Local Importance(LI). To evaluate their effectiveness, we conduct a comprehensive ablation study using a shared adapter-based method ($\mathcal{A}^t$) as the baseline (Idx1). The experiments are carried out across 10 tasks on Split-ImageNetA and Split-ImageNetR, as detailed in Table \ref{tab:GI_and_LI}. Specifically, Idx2 extends the baseline by incorporating GI, while Idx3 introduces LI independently. Idx4 further combines both components, facilitating the evaluation of joint effects.
The experimental results demonstrate that incorporating either GI or LI into the baseline improves performance across both benchmarks. Specifically, adding GI alone (Idx2) leads to 1.05\% and 0.23\% improvement of $A_{Last}$ in Split-ImageNetA and Split-ImageNetR,  0.88\% and 0.65\% improvement of $A_{Avg}$ in Split-ImageNetA and Split-ImageNetR. Similarly, integrating LI (Idx3) also contributes positively: 0.94\% and 0.65\% improvement of $A_{Last}$, 0.83\% and 1.02\% improvement of $A_{Avg}$ in Split-ImageNetA and Split-ImageNetR. Since GI and LI can quantify the sensitivity of parameters to past tasks from different aspects, the combined integration of GI and LI (Idx4) achieves the highest performance across all evaluated metrics, with $A_{Last}$ reaching 59.14\% on Split-ImageNetA and 78.75\% on Split-ImageNetR, while $A_{Avg}$ improving to 67.98\% and 83.92\%, respectively.
These results underscore the combined roles of GI and LI in enhancing model performance. Specifically, GI quantifies the aggregate contribution of all adapter
modules by formulating an optimization objective of the final output feature, LI captures the specific contribution of individual adapter modules on the output. Their joint implementation effectively combines the shared and specific contributions of different adapter modules to output, more comprehensively quantifying the sensitivity of the model to past tasks. Therefore, GI and LI allow Importance-aware Parameter Regularization (IPR) to strictly constrain parameters that are important to old tasks to reduce forgetting, while relaxing the constraints of relatively unimportant parameters to ensure the plasticity of the model.\\

\noindent\textbf{Further analysis of IPR's importance.}
The IPR module introduces an innovative method for quantifying parameter importance and implementing regularization by:
1) Quantifying the aggregate contribution of all adapter
modules global importance by formulating an optimization objective that maximizes the classification score while minimizing variance across different channels of the final output feature; 2) Capturing the specific contribution of individual adapter modules via local importance that analyzes the impact of adding perturbations to specific modules on the output; and 3) Applying an importance-aware parameter regularization strategy to mitigate catastrophic forgetting while preserving the model's capacity for acquiring new knowledge.\par
Existed uniform Parameter Regularization (PR) methods impose uniform constraints on all weights, thereby sacrificing the model plasticity to a large extent. In contrast, IPR applies parameter-specific constraints based on parameter importance, selectively protecting critical parameters while allowing less important ones to adapt. Moreover, the Amplitude-based Parameter Regularization (APR) methods neglect feature interaction effects, whereas IPR incorporates propagation sensitivity analysis by quantifying the actual impact of parameters on the output. Furthermore, the Fisher matrix–based Parameter Regularization (FPR) relies on a diagonal approximation of the Fisher matrix, which disregards parameter correlations and assumes that parameters follow a Gaussian distribution—a simplification that deviates from the true posterior and calculating the fisher information matrix requires computationally expensive backpropagation. In contrast, IPR directly models the parameter–performance relationship through feature statistics, thereby avoiding distributional assumptions. By leveraging forward propagation to compute parameter importance, IPR reduces computational complexity while explicitly capturing cross-layer dependencies, effectively accounting for parameter correlations and improving overall regularization efficacy. \par
To further validate the rationale and effectiveness of IPR, we design and compare various parameter regularization methods within a shared adapter-tuning ($\mathcal{A}^t$), with results shown in Table.\ref{tab: PR vs Fisher vs ABS vs IPR}. The evaluated methods include uniform Parameter Regularization (PR, Idx2), Amplitude-based Parameter Regularization (APR, Idx3), Fisher matrix-aware Parameter Regularization (FPR, Idx4) and our proposed Importance-aware Parameter Regularization (IPR, Idx5). Specifically, PR imposes uniform constraints on all parameters, APR utilizes the magnitude of the parameters from the previous model as regularization weights, and FPR employs the fisher matrix computed from the gradients of the previous model to determine the regularization strength. In comparison, our proposed IPR method dynamically assigns parameter-specific constraints based on their computed importance, ensuring a more effective and adaptive regularization strategy. Experimental results demonstrate that IPR consistently outperforms all alternative regularization methods, highlighting the superiority of its importance estimation mechanism.\par
\begin{figure}[htbp]
\centering
\includegraphics[width=\linewidth]{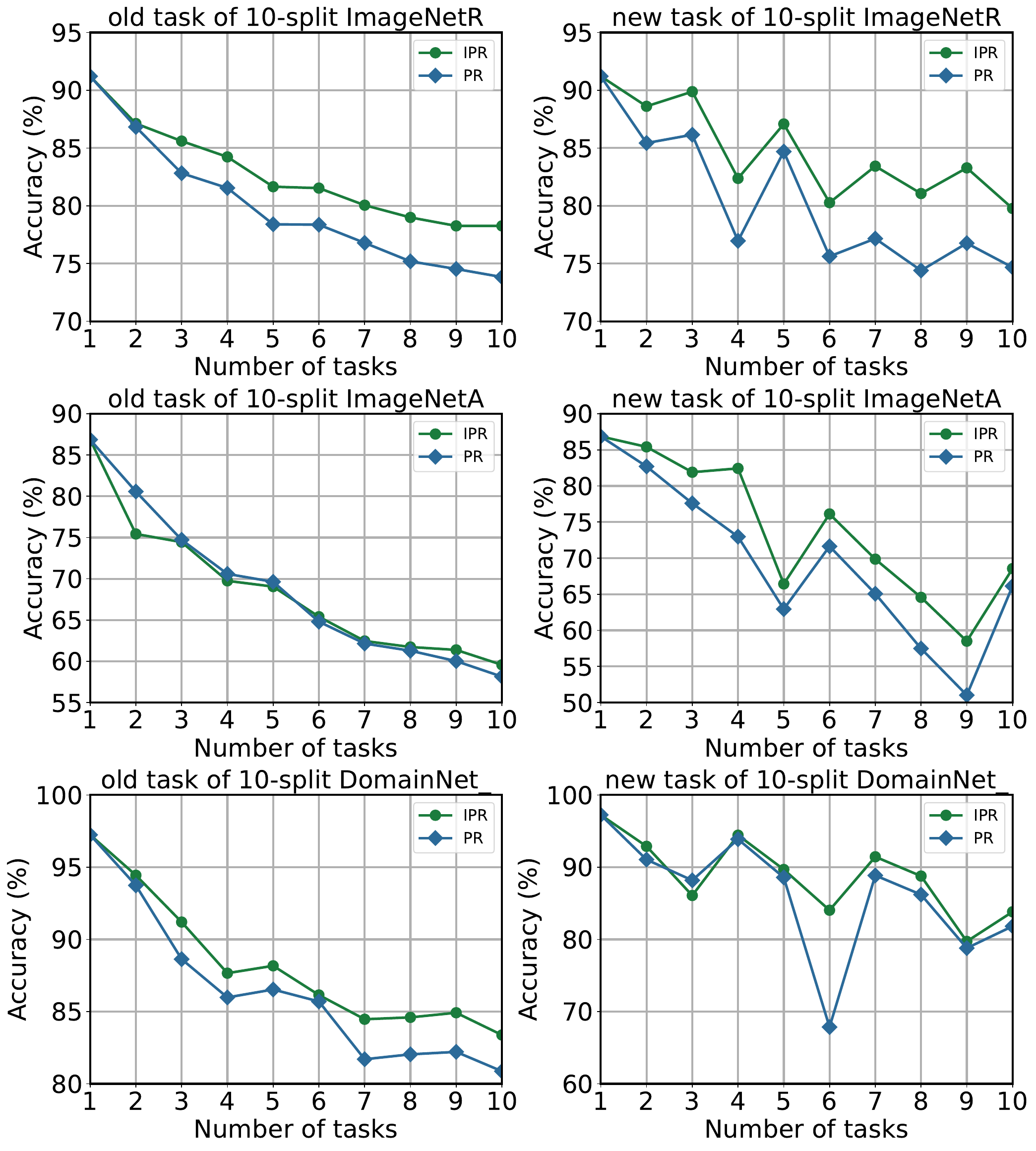}
\caption{old and new task accuracy for incremental sessions on three CIL benchmarks under the method of IPR and PR.}
\label{IPR_vs_PR}
\end{figure}

Fig \ref{IPR_vs_PR} provides a detailed comparison of the performance metrics for both old and new tasks within a 10-task continual learning setting. As Global Importance (GI) quantifies the aggregate contribution of all adapter modules and Local Importance (LI) captures the specific contribution of individual adapter modules, the proposed Importance-aware Parameter Regularization (IPR) method selectively strengthens constraints on parameters that are crucial for old tasks while allowing other parameters greater flexibility to learn new tasks. Compared with uniform Parameter Regularization (PR), IPR demonstrates superior knowledge retention for old tasks while simultaneously enhancing the model's adaptability to new tasks. 
As a result, models trained with IPR consistently achieve higher accuracy on both old and new tasks. These results demonstrate that IPR is significantly more effective in mitigating catastrophic forgetting while facilitating the acquisition of new tasks, thereby providing a more robust and adaptive solution to continual learning challenges. \par
\noindent\textbf{Further analysis of TSDC’s trainable Drift}
Since the model has semantic drift when training for new tasks, we need to train a unified classifier. The Without-Semantic Drift (Wo-SD) method constructs a Gaussian sampling model by preserving old class prototypes. It then samples and simulates data from previous tasks to approximate joint training of the classifier. Building upon this, the Static Semantic Drift (Static) method estimates feature space drift during the learning of new tasks. It compensates for this drift by new class data after training convergence, under the assumption that the error is acceptable when the drift is small enough.
To overcome the limitation of assuming small drift, the Trainable Semantic Drift Compensation (TSDC) method actively minimizes the gap between estimated drift and the zero subspace during training. This refinement enhances drift estimation accuracy, thereby improving continual learning performance. As shown in Table.\ref{tab: Diff_SD}, sampling without drift compensation leads to suboptimal performance and Static Semantic Drift method with larger drift value cannot achieve optimal performance. In contrast, the proposed trainable drift update strategy effectively mitigates errors induced by substantial feature drift, ultimately achieving superior performance.\par

\begin{table}[t]
\centering
\renewcommand{\arraystretch}{1.8} 
\caption{Results of existing methods under other pre-trained model. DINO-1k denotes that the frozen backbone is pre-trained on ImageNet-1k through DINO pre-training algorithms. We report the averaged results over 3 random number seeds. The highest results are in bold and the second highest results are underlined.}\label{tab: dino}
\setlength{\tabcolsep}{5pt} 
\scalebox{0.78}{
\begin{tabular}{c|ccc}
\hline
PTM & Method & $A_{Last}$ $\uparrow$ &$A_{Avg}$ $\uparrow$\\
\hline
\multirow{9}*{DINO-1k}  &L2P \cite{wang2022learning} & 56.71\tsb{\(\pm\)0.12} & 63.59\tsb{\(\pm\)0.21}\\
{} & DualPrompt \cite{wang2022dualprompt} & 60.23\tsb{\(\pm\)0.42} & 66.57\tsb{\(\pm\)0.25} \\
{} & CODAPrompt \cite{CODAPromptCOntinual2023Smith} & 64.02\tsb{\(\pm\)0.68} &71.50\tsb{\(\pm\)0.42} \\
{} & C-LoRA \cite{smith2023continual} & 63.07\tsb{\(\pm\)0.36} & 68.09\tsb{\(\pm\)0.41} \\
{} & LAE \cite{gao2023unified} & 61.03\tsb{\(\pm\)0.27} & 69.89\tsb{\(\pm\)0.15} \\
{} & HiDe-Prompt \cite{wang2024hierarchical} & 68.11\tsb{\(\pm\)0.18} & 71.70\tsb{\(\pm\)0.01} \\
{} & InfLoRA \cite{liang2024inflora} & 68.31\tsb{\(\pm\)0.28} & 76.15\tsb{\(\pm\)0.05} \\
{} & VPS-NSP \cite{lu2024visual} & \underline{68.96}\tsb{\(\pm\)0.94} & \underline{76.22}\tsb{\(\pm\)0.56}\\
\hline
{} & \textbf{EKPC (Ours)} &\textbf{72.67}\tsb{\(\pm\)0.35} & \textbf{78.26}\tsb{\(\pm\)0.78}\\
\hline
\end{tabular}}
\label{performance}
\end{table}
\subsection{Further Analysis}
\noindent\textbf{Experiments under other pre-trained model.} Following \cite{liang2024inflora, wang2024hierarchical}, we conduct experiments on 10s-ImageNetR using a ViT-B/16 model pre-trained with the DINO self-supervised method \cite{caron2021emerging}. As presented in Table.\ref{tab: dino}, the results show a performance decline when using a self-supervised pre-trained backbone compared to a backbone pre-trained in a supervised manner. Nevertheless, our proposed method consistently outperforms other PEFT-based approaches. These results exhibit the adaptability of our approach across models trained with different pre-training paradigms, highlighting its robustness and generalization.\\
\noindent\textbf{Ablation Study for decision boundary confusion.}
To demonstrate the effectiveness of our method in mitigating the confusion of decision boundaries across tasks, Table.\ref{tab: task_results} presents the task-by-task accuracy of task ID recognition across multiple datasets, including ImageNetA, ImageNetR, and DomainNet. The results highlight the impact of progressively integrating the proposed modules (IPR and TSDC) into the baseline model. Averaged over three random seeds, the results illustrate the contribution of these modules in enhancing recognition accuracy across successive sessions.\par
As shown in Table.\ref{tab: task_results}, the baseline model (with $\mathcal{A}^t$ only) consistently exhibits performance degradation over sequential learning sessions. Specifically, under the session 10 setting, its accuracy drops from an initial 100\% to 58.31\% on ImageNetA, 79.30\% on ImageNetR, and 83.40\% on DomainNet by the final session.
Since the IPR module primarily retains class-specific information, its effectiveness in mitigating cross-task decision boundary confusion is relatively limited, yielding a modest improvement of 1.24\%. In contrast, the TSDC module significantly enhances task ID recognition accuracy by 5.02\%, emphasizing the synergy between the two modules.
The final proposed framework ($\mathcal{A}^t$+IPR+TSDC) achieves optimal performance across all datasets, with an overall improvement of 5.65\%, i.e., 67.32\% on ImageNetA, 82.30\% on ImageNetR, and 88.35\% on DomainNet accuracy in final sessions. These results demonstrate the effectiveness of the proposed method in improving task ID recognition and improving the model’s ability to mitigate cross-task decision boundary confusion.\\
\begin{table*}[t]
\renewcommand{\arraystretch}{1.8} 
\centering
\caption{Task-by-task accuracy of task ID recognition for different Datasets. We add our modules one by one based on the baseline and report the averaged results over 3 random number seeds. The highest results are in bold.}\label{tab: task_results}
\setlength{\tabcolsep}{2pt} 
\begin{adjustbox}{max width=1\textwidth}
\begin{tabular}{c|c@{\hspace{5pt}}c@{\hspace{5pt}}c@{\hspace{5pt}}ccccccccccc}
\toprule
\multirow{2}*{Datasets} &
\multirow{2}*{Idx} & 
\multirow{2}*{$\mathcal{A}^t$} &
\multirow{2}*{IPR} &  
\multirow{2}*{TSDC} &
\multirow{2}*{$Ses.1$} &
\multirow{2}*{$Ses.2$} &
\multirow{2}*{$Ses.3$} &
\multirow{2}*{$Ses.4$} &
\multirow{2}*{$Ses.5$} &
\multirow{2}*{$Ses.6$} &
\multirow{2}*{$Ses.7$} &
\multirow{2}*{$Ses.8$} &
\multirow{2}*{$Ses.9$} &
\multirow{2}*{$Ses.10$} \\
& & & & & & & & & & & & & &\\
\hline
\multirow{4}*{ImageNetA} & 1 & \checkmark & -- & -- & 100.00\tsb{\(\pm\)0.00} & 84.54\tsb{\(\pm\)0.44} & 78.54\tsb{\(\pm\)0.42} & 72.81\tsb{\(\pm\)1.90} & 69.40\tsb{\(\pm\)2.37} & 66.77\tsb{\(\pm\)1.34}  & 62.28\tsb{\(\pm\)1.99}  & 61.45\tsb{\(\pm\)1.29} & 60.30\tsb{\(\pm\)1.33} & 58.31\tsb{\(\pm\)1.89} \\
{} & 2 & \checkmark & \checkmark & -- & 100.00\tsb{\(\pm\)0.00} & 86.57\tsb{\(\pm\)1.20} & 77.67\tsb{\(\pm\)0.16} & 70.57\tsb{\(\pm\)0.91} & 67.44\tsb{\(\pm\)1.34} & 63.61\tsb{\(\pm\)0.99}  & 63.61\tsb{\(\pm\)0.99}  & 62.66\tsb{\(\pm\)0.67} & 60.56\tsb{\(\pm\)0.45} & 60.24\tsb{\(\pm\)0.92} \\
{} & 3 & \checkmark & -- & \checkmark & 100.00\tsb{\(\pm\)0.00} & 88.63\tsb{\(\pm\)1.58} & 82.28\tsb{\(\pm\)0.63} & 78.81\tsb{\(\pm\)1.57} & 76.76\tsb{\(\pm\)0.90} & 73.72\tsb{\(\pm\)0.52}  & \textbf{71.66}\tsb{\(\pm\)0.98}  & \textbf{69.99}\tsb{\(\pm\)0.72} & \textbf{68.02}\tsb{\(\pm\)1.21} & 66.98\tsb{\(\pm\)1.41} \\
{} & 4 & \checkmark & \checkmark & \checkmark & \textbf{100.00}\tsb{\(\pm\)0.00} & \textbf{89.34}\tsb{\(\pm\)1.19} & \textbf{83.44}\tsb{\(\pm\)0.62} & \textbf{79.50}\tsb{\(\pm\)1.70} & \textbf{76.93}\tsb{\(\pm\)0.65} & \textbf{74.18}\tsb{\(\pm\)0.64}  & 71.34\tsb{\(\pm\)0.47}  & 69.58\tsb{\(\pm\)0.51} & 67.62\tsb{\(\pm\)0.88} & \textbf{67.32}\tsb{\(\pm\)1.75} \\
\hline
\multirow{4}*{ImageNetR} &1 & \checkmark & -- & -- & 100.00\tsb{\(\pm\)0.00} & 93.00\tsb{\(\pm\)0.80} & 89.67\tsb{\(\pm\)0.53} & 86.70\tsb{\(\pm\)0.41} & 84.42\tsb{\(\pm\)0.77} & 83.35\tsb{\(\pm\)0.67}  & 82.22\tsb{\(\pm\)0.93}  & 81.07\tsb{\(\pm\)0.95} & 80.12\tsb{\(\pm\)0.95} & 79.30\tsb{\(\pm\)0.37} \\
{} & 2 & \checkmark & \checkmark & -- & 100.00\tsb{\(\pm\)0.00} & 93.51\tsb{\(\pm\)0.83} & 90.69\tsb{\(\pm\)0.36} & 87.49\tsb{\(\pm\)0.61} & 85.36\tsb{\(\pm\)0.36} & 83.92\tsb{\(\pm\)0.48}  & 83.07\tsb{\(\pm\)0.68}  & 81.70\tsb{\(\pm\)0.97} & 80.91\tsb{\(\pm\)0.75} & 80.20\tsb{\(\pm\)0.42} \\
{} & 3 & \checkmark & -- & \checkmark & 100.00\tsb{\(\pm\)0.00} & 94.33\tsb{\(\pm\)1.43} & \textbf{91.34}\tsb{\(\pm\)0.33} & \textbf{88.79}\tsb{\(\pm\)0.24} & \textbf{86.67}\tsb{\(\pm\)0.32} & 85.38\tsb{\(\pm\)0.42}  & 84.24\tsb{\(\pm\)0.53}  & 83.34\tsb{\(\pm\)0.51} & 82.27\tsb{\(\pm\)0.45} & 81.75\tsb{\(\pm\)0.72} \\
{} & 4 & \checkmark & \checkmark & \checkmark & \textbf{100.00}\tsb{\(\pm\)0.00} & \textbf{94.44}\tsb{\(\pm\)0.25} & 91.25\tsb{\(\pm\)0.47} & 88.63\tsb{\(\pm\)0.33} & 86.65\tsb{\(\pm\)0.25} & \textbf{85.49}\tsb{\(\pm\)0.24}  & \textbf{84.75}\tsb{\(\pm\)0.53}  & \textbf{83.70}\tsb{\(\pm\)0.41} & \textbf{82.98}\tsb{\(\pm\)0.35} & \textbf{82.30}\tsb{\(\pm\)0.27} \\
\hline
\multirow{4}*{DomainNet} & 1 & \checkmark & -- & -- & 100.00\tsb{\(\pm\)0.00} & 95.50\tsb{\(\pm\)0.63} & 92.31\tsb{\(\pm\)1.76} & 90.24\tsb{\(\pm\)1.06} & 88.69\tsb{\(\pm\)0.45} & 86.65\tsb{\(\pm\)0.72}  & 85.84\tsb{\(\pm\)0.26}  & 85.16\tsb{\(\pm\)0.44} & 84.21\tsb{\(\pm\)0.41} & 83.40\tsb{\(\pm\)0.31} \\
{} & 2 & \checkmark & \checkmark & -- & 100.00\tsb{\(\pm\)0.00} & 95.72\tsb{\(\pm\)0.94} & 93.02\tsb{\(\pm\)1.80} & 91.02\tsb{\(\pm\)1.28} & 89.55\tsb{\(\pm\)0.41} & 87.54\tsb{\(\pm\)0.76}  & 86.66\tsb{\(\pm\)0.36}  & 85.95\tsb{\(\pm\)0.72} & 85.00\tsb{\(\pm\)0.43} & 84.28\tsb{\(\pm\)0.40} \\
{} & 3 & \checkmark & -- & \checkmark & 100.00\tsb{\(\pm\)0.00} & \textbf{97.19}\tsb{\(\pm\)0.25} & \textbf{95.03}\tsb{\(\pm\)1.24} & 93.79\tsb{\(\pm\)0.61} & 92.46\tsb{\(\pm\)0.20} & 90.96\tsb{\(\pm\)0.74}  & 89.81\tsb{\(\pm\)0.81}  & 89.18\tsb{\(\pm\)0.33} & 88.31\tsb{\(\pm\)0.28} & 87.35\tsb{\(\pm\)0.64} \\
{} & 4 & \checkmark & \checkmark & \checkmark & \textbf{100.00}\tsb{\(\pm\)0.00} & 97.15\tsb{\(\pm\)0.27} & 95.01\tsb{\(\pm\)1.23} & \textbf{93.94}\tsb{\(\pm\)0.68} & \textbf{92.51}\tsb{\(\pm\)0.22} & \textbf{91.23}\tsb{\(\pm\)0.76}  & \textbf{90.25}\tsb{\(\pm\)0.78}  & \textbf{89.78}\tsb{\(\pm\)0.38} & \textbf{89.13}\tsb{\(\pm\)0.32} & \textbf{88.35}\tsb{\(\pm\)0.23} \\
\bottomrule
\end{tabular}
\end{adjustbox}
\end{table*}
\begin{table}[t]
\centering
\caption{hyper-parameter analysis of \( w_1 \) and \( w_2 \) on 10-tasks of S-ImageNetR. We report the averaged results over 3 random number seeds. The highest results are in bold.}\label{tab: w1_w2}
\begin{adjustbox}{max width=1.0\textwidth}
\begin{tabular}{c|ccccc}
\toprule
\multirow{2}*{hyper-parameter} & \multirow{2}*{Value} & 
\multirow{2}*{$A_{Last}$ $\uparrow$} &
\multirow{2}*{$A_{Avg}$ $\uparrow$} \\

& & & \\
\hline
\multirow{5}*{\( w_1 \)} & 0.1 & 80.45\tsb{\(\pm\)0.13} & 84.92\tsb{\(\pm\)0.46} \\
{} & 1 & \textbf{80.60}\tsb{\(\pm\)0.08} & 84.92\tsb{\(\pm\)0.48} \\
{}  & 2 & 80.33\tsb{\(\pm\)0.15} & 84.78\tsb{\(\pm\)0.36} \\
{}  & 3 & 80.27\tsb{\(\pm\)0.16} & \textbf{84.96}\tsb{\(\pm\)0.11} \\
{}  & 4 & 80.34\tsb{\(\pm\)0.21} & 84.72\tsb{\(\pm\)0.37} \\
\hline
\multirow{5}*{\( w_2 \)} & 0.1 & 80.28\tsb{\(\pm\)0.15} & 84.83\tsb{\(\pm\)0.37} \\
{} & 1 & \textbf{80.60}\tsb{\(\pm\)0.08} & \textbf{84.92}\tsb{\(\pm\)0.48} \\
{} & 2 & 80.48\tsb{\(\pm\)0.20} & 84.84\tsb{\(\pm\)0.39} \\
{} & 3 & 80.52\tsb{\(\pm\)0.19} & 84.82\tsb{\(\pm\)0.38} \\
{} & 4 & 80.47\tsb{\(\pm\)0.23} & 84.79\tsb{\(\pm\)0.39} \\
\bottomrule
\end{tabular}
\end{adjustbox}
\vspace{-4mm}
\end{table} 


\begin{figure}[htbp]
\centering
\includegraphics[width=\linewidth]{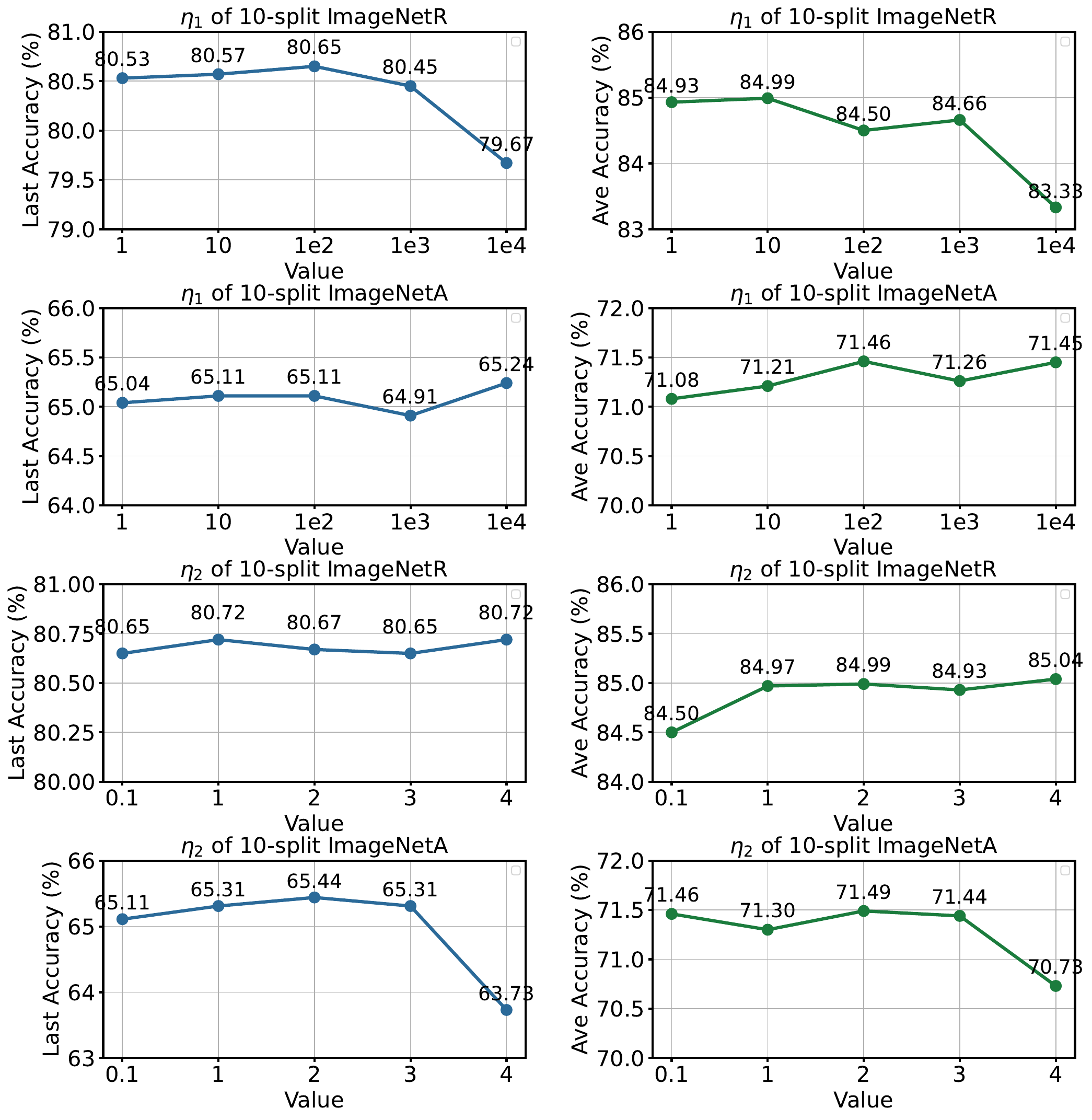}
\caption{hyper-parameter analysis of $\eta_1$ and $\eta_2$ on 10-tasks of S-ImageNetR and S-ImageNetA.}
\label{eta_1_eta_2}
\end{figure}
\noindent\textbf{Hyper-parameter Analysis.}
To analyze the sensitivity of the model to hyper-parameters, we define specific variation intervals for different hyper-parameters. Fig.\ref{eta_1_eta_2} and Table.\ref{tab: w1_w2} show the impact of these variations on model performance.\\
\noindent \textbf{Analysis of hyper-parameters $w_1$ and $w_2$}:
Table \ref{tab: w1_w2} evaluates the sensitivity of the weighting parameters $w_1$ and $w_2$ in the 10-task Split-ImageNetR benchmark, with two key observations: 1). Parameter Robustness: The experimental results demonstrate that the proposed method exhibits high robustness to variations in $w_1$ and $w_2$, with performance fluctuations remaining within 0.33\% when adjusting $w_1$ or $w_2$ in the range of [0.1, 4]. This indicates that the method is relatively insensitive to moderate changes in these hyperparameters. 2). Optimality Principle: The results further reveal that the balanced configuration ($w_1=1$, $w_2=1$) maximizes both the performance preservation of the last task ($A_{Last}=80.60\%$) and the cumulative knowledge retention of all tasks ( $A_{Avg}=84.92\%$), indicating the default weights optimally balance stability-plasticity trade-off. The findings confirm that the proposed approach has low dependence on hyperparameter tuning, with performance variations remaining below 1\% across the tested parameter ranges. This property is particularly advantageous in continual learning settings, as it minimizes the need for extensive hyperparameter tuning during deployment, thereby improving practicality and scalability.\\
\noindent \textbf{Analysis of hyper-parameters $\eta_1$ and $\eta_2$}:
The Fig.\ref{eta_1_eta_2} illustrates the influence of hyper-parameters $\eta_1$ and $\eta_2$ on both the Last Accuracy and Average Accuracy, evaluated on the 10-split ImageNetR and ImageNetA benchmarks. The results lead to the following key observations: 1). Parameter Robustness: The experimental results demonstrate the robustness of our approach, with performance variations remaining below 1.60\% when adjusting $\eta_1$ within the range of $1 \to 10^4$ and $\eta_2$ within $0.1 \to 4$. 2). Optimality Principle: The configuration $\eta_1 = 100$ and $\eta_2 = 1$ optimally balances both the preservation of last-task performance ($A_{Last} = 80.65\%$ on ImageNetR and $A_{Last} = 65.11\%$ on ImageNetA) and the retention of cumulative knowledge across all tasks ($A_{Avg} = 84.50\%$ on ImageNetR and $A_{Avg} = 71.46\%$ on ImageNetA). The findings highlight that our method exhibits minimal dependence on hyperparameter tuning, with performance variation constrained within 2\% across the tested ranges. This property significantly reduces the burden of hyperparameter optimization in continual learning deployments.

\begin{table}
\centering
\caption{Average forgetting between SSIAT and EKPC on five CIL benchmarks with 10/20 incremental sessions. The results of the lowest forgetfulness are in bold}
\setlength{\tabcolsep}{10pt} 
\scalebox{1}{ 
\begin{tabular}{c|ccccc}
\toprule
 \multirow{2}*{Datasets} &\multicolumn{1}{c}{SSIAT} &\multicolumn{1}{c}{EKPC}  \\
 & $AF\downarrow$& $AF\downarrow$\\
\hline
\multirow{1}*{10s-CIFAR100} & 4.67\tsb{\(\pm\)0.63} & \textbf{3.26\tsb{\(\pm\)0.32}} \\
\multirow{1}*{10s-CUB200} & 4.71\tsb{\(\pm\)0.46} & \textbf{3.78\tsb{\(\pm\)0.16}} \\
\multirow{1}*{10s-ImageNetA} & 15.93\tsb{\(\pm\)0.41} & \textbf{10.45\tsb{\(\pm\)0.51}} \\
\multirow{1}*{10s-ImageNetR} & 6.35\tsb{\(\pm\)0.54} & \textbf{5.48\tsb{\(\pm\)0.12}} \\
\multirow{1}*{10s-DomainNet} & 7.72\tsb{\(\pm\)1.04} & \textbf{4.17\tsb{\(\pm\)0.10}} \\
\multirow{1}*{20s-ImageNetR} & 6.06\tsb{\(\pm\)0.18} & \textbf{5.91\tsb{\(\pm\)0.12}} \\
\multirow{1}*{20s-ImageNetA} & 18.73\tsb{\(\pm\)2.31} & \textbf{14.26\tsb{\(\pm\)1.24}} \\
\bottomrule
\end{tabular}}
\label{tab: forgetting}
\end{table}
\noindent\textbf{Analysis of Forgetting}: To evaluate the robustness of our method against catastrophic forgetting, we quantify knowledge retention using the Average Forgetting (AF) metric $AF=\frac{1}{T-1} \sum^{T-1}_{t=1}\left( Acc^{initial}_{t} - Acc^{final}_{t} \right)$, where $Acc^{initial}_{t}$ and $Acc^{final}_{t}$ denote the accuracy on task t immediately after its training and at the end of the entire training sequence, respectively. A lower AF value indicates a stronger resistance to forgetting. To further validate the effectiveness of our approach, we compare the AF metric with the SOTA method (SSIAT) on five benchmark datasets. As shown in Table.\ref{tab: forgetting}, our method consistently achieves lower AF values than SSIAT, demonstrating its superior ability to mitigate catastrophic forgetting.\par
\section{Conclusion}
In this paper, we propose the Elastic Knowledge Preservation and Compensation (EKPC) method for class-incremental learning. Unlike existing parameter-efficient fine-tuning (PEFT) methods that either introduce excessive additional parameters or impose overly rigid regularization constraints, EKPC achieves a better balance between knowledge preservation and model plasticity. Central to EKPC is the Importance-aware Parameter Regularization (IPR) method, which quantifies the sensitivity of network parameters to prior tasks through a novel parameter-importance algorithm coupled with elastic regularization. This strategy selectively constrains parameter updates based on their importance to previous tasks, ensuring stable knowledge retention without sacrificing plasticity. In addition, we mitigate the challenge of decision boundary confusion in classifiers by proposing the Trainable Semantic Drift Compensation (TSDC) method. Training a unified classifier by actively compensating semantic drift for sample space, TSDC refines the decision boundaries, ensuring robust classification performance as new classes are introduced. Future work will focus on extending EKPC to more complex continual learning scenarios, such as online setting.\\
\bibliographystyle{spbasic}
\bibliography{sn-bibliography}    

%
%

\end{sloppypar}
\end{document}